\documentclass[10pt,twocolumn,letterpaper]{article}

\usepackage{cvpr}  

\usepackage[dvipsnames]{xcolor}

\newcommand{\bI}{\mathbf{I}}

\newcommand{\bS}{\mathbf{S}}
\newcommand{\bt}{\mathbf{t}}

\newcommand{\figref}[1]{Figure~\ref{#1}}
\newcommand{\secref}[1]{Section~\ref{#1}}

\newcommand{\tabref}[1]{Table~\ref{#1}}

\def\mc{\mathcal}

\makeatletter
\DeclareRobustCommand\onedot{\futurelet\@let@token\@onedot}
\def\@onedot{\ifx\@let@token.\else.\null\fi\xspace}
\def\eg{e.g\onedot} 
\def\ie{i.e\onedot}

\def\etal{et~al\onedot}

\makeatother

\definecolor{yellow}{rgb}{1, 1, 0.7}
\definecolor{orange}{rgb}{1, 0.85, 0.7}
\definecolor{tablered}{rgb}{1, 0.7, 0.7}
\definecolor{red}{rgb}{1, 0, 0}
\newcommand{\best}{\cellcolor{tablered}}
\newcommand{\sbest}{\cellcolor{orange}}
\newcommand{\tbest}{\cellcolor{yellow}}

\definecolor{wincolor}{rgb}{0.85, 0.0, 0.0}

\definecolor{darkyellow}{rgb}{0.8, 0.8, 0.5}
\definecolor{darkred}{rgb}{0.7, 0.3, 0.3}
\definecolor{darkgreen}{rgb}{0.3, 0.7, 0.3}
\definecolor{blue}{rgb}{0.251, 0.498, 0.824}
\definecolor{green}{rgb}{0, 1.0, 0}
\definecolor{pink}{rgb}{1, 0.4, 0.7}

\definecolor{realred}{rgb}{0.95, 0.1, 0.0}
\newcommand{\red}[1]{{\color{realred}#1}}

\newcommand{\boldparagraph}[1]{\vspace{0.1cm}\noindent{\bf #1:}}

\definecolor{cvprblue}{rgb}{0.21,0.49,0.74}
\usepackage[pagebackref,breaklinks,colorlinks,citecolor=cvprblue]{hyperref}
\usepackage{pifont}
\usepackage{color}
\usepackage{xcolor}
\usepackage{nicefrac}       
\usepackage{multirow}
\usepackage{multicol}
\usepackage{listings}
\usepackage{dsfont}

\usepackage{tikzducks}
\usepackage{xfp,graphicx}

\usepackage{colortbl}
\usepackage{makecell}
\usepackage[percent]{overpic}
\usepackage{tikz}
\usepackage[accsupp]{axessibility} 

\newcommand{\boldstartspace}[1]{\vspace{0.1in}\noindent\textbf{#1}}

\def\paperID{4321} 
\def\confName{CVPR}
\def\confYear{2025}

\title{GenFusion: Closing the Loop between Reconstruction and Generation via Videos}

\author{\fontsize{11.25pt}{\baselineskip}\selectfont Sibo Wu\textsuperscript{1,2}~~~Congrong Xu\textsuperscript{1,3}~~~Binbin Huang\textsuperscript{4}~~~Andreas Geiger\textsuperscript{5}~~~Anpei Chen\textsuperscript{1,5,\textdagger}\\[2mm]
\fontsize{11.25pt}{\baselineskip}\selectfont \textsuperscript{1}Westlake University~~~~\textsuperscript{2}Technical University of Munich~~~~\textsuperscript{3}ShanghaiTech University\\[0.5mm]
\fontsize{11.25pt}{\baselineskip}\selectfont\textsuperscript{4}The University of Hong Kong~~~~\textsuperscript{5}University of Tübingen, Tübingen AI Center~~~~\textsuperscript{\textdagger}Corresponding author
}

\begin{document}


\newcommand\getduck{\fpeval{randint(1,100)}}
\renewcommand{\thefootnote}{\fnsymbol{footnote}}

\renewcommand{\red}{}


\twocolumn[{%
\renewcommand\twocolumn[1][]{#1}%
\maketitle
\begin{center}
\vspace{-2mm}
    \centering
    \includegraphics[width=\textwidth]{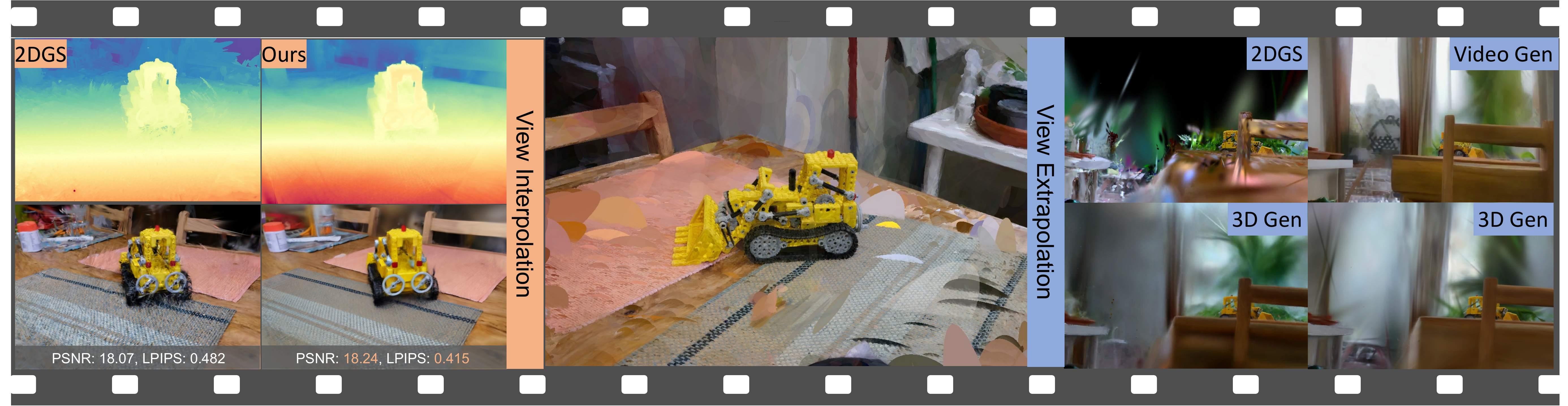}
    \captionof{figure}{\textbf{GenFusion} introduces a reconstruction-driven generative model to enable artifact-free 3D asset generation and view synthesis for both view interpolation and extrapolation.}
    \label{fig:teaser}
\end{center}
}]


\begin{abstract}
\vspace{-0.1in}
Recently, 3D reconstruction and generation have demonstrated impressive novel view synthesis results, achieving high fidelity and efficiency. 
However, a notable conditioning gap can be observed between these two fields, \eg, scalable 3D scene reconstruction often requires densely captured views, whereas 3D generation typically relies on a single or no input view, which significantly limits their applications.
We found that the source of this phenomenon lies in the misalignment between 3D constraints and generative priors.
To address this problem, we propose a reconstruction-driven video diffusion model that learns to condition video frames on artifact-prone RGB-D renderings.
%
%
Moreover, we propose a cyclical fusion pipeline that iteratively adds restoration frames from the generative model to the training set,
enabling progressive expansion and addressing the viewpoint saturation limitations seen in previous reconstruction and generation pipelines.
Our evaluation, including view synthesis from sparse view and masked input, validates the effectiveness of our approach. More details at \url{https://genfusion.sibowu.com}.
\vspace{-0.1in}
\end{abstract}

\section{Introduction}
Generating 3D assets is a fundamental task in computer vision and computer graphics, with broad applications in AR/VR, autonomous driving and robotics. 
Recent advances in Neural Radiance Fields (NeRF~\cite{Mildenhall2020ECCV}, Mildenhall 
 \etal in 2020) and Gaussian Splatting (GS~\cite{Kerbl2023TOG}, Kerbl \etal in 2023) have enabled high-fidelity 3D scene reconstruction and novel view synthesis.
They employ MLP or Gaussian primitives to represent scenes and optimize 3D representation through photometric loss.
However, this line of work inherits a key limitation: faithful reconstruction relies on abundant viewpoint coverage; under-observed regions or viewpoints may lead to significant artifacts or missing content. 

This is primarily because reconstructing NeRFs or GSs from multi-view images is inherently underconstrained, as an infinite number of photo-consistent explanations may exist for the input images~\cite{kaizhang2020nerfplusplus,Barbara2022CVPR}. Consequently, reconstruction models tend to generate ``floaters" or ``background collapse" artifacts to fake view-dependent effects, even when supplied with dense and well-captured high-quality images~\cite{Nerfbusters2023}. This observation has motivated a series of regularization techniques to constrain neural field training, including sparsity regularizers~\cite{Hedman2021ICCV, Yu2021ICCV}, smoothness losses~\cite{Niemeyer2022CVPR, Yang2023CVPR, Zhang2024CVPR}
and monocular geometric cues~\cite{Yu2022NEURIPS}. For example, ReconFusion~\cite{Rundi2024CVPR} regularizes a NeRF-based 3D reconstruction pipeline by introducing a sample loss between novel random camera poses and images predicted by a PixelNeRF-style image diffusion model, yielding significant performance improvements over previous NeRF reconstruction methods in sparse setting.
In contrast, feedforward reconstruction methods~\cite{Alex2021CVPR, Chen2021ICCVb, David2024CVPR, Yuedong2024MVSplat, LaRa, Wangbo2024ViewCrafter} learn inductive biases directly from the dataset. While recent advances enable 3D reconstruction from as few as a single image, existing feedforward reconstruction methods exhibit performance saturation when processing more than 4-8 images. This limitation arises primarily from the architectural constraints of conventional feedforward networks in effectively aggregating and utilizing information from multiple viewpoints.

Meanwhile, generative methods have demonstrated the potential of obtaining 3D assets without multi-view capture. Leveraging large-scale datasets and scalable architectures, models like Stable Diffusion (SD) have achieved remarkable progress in image and video generation~\cite{DDPM,SVD, guo2024i2v}. Recent work has applied these approaches to generate 3D assets. For example, DreamFusion~\cite{DreamFusion} introduces Score Distillation Sampling (SDS) to perform text-to-3D synthesis using a pre-trained 2D text-to-image diffusion model, while another line of research~\cite{li2022_infinite_nature_zero, cai2022diffdreamer,yu2024wonderjourney} explores single-view scene extrapolation by progressively in/outpainting layered depth image.

Despite these advances in 3D reconstruction and generation, a notable conditioning gap remains: scalable 3D reconstruction typically requires dense view coverage, whereas generation methods often operate with single or even no input view.
Our paper explores how 3D reconstruction and generation can complement each other in a scalable manner, relaxing the constraints on the number of input views. 

\begin{figure*}
    \centering
    \includegraphics[width=0.95\textwidth]{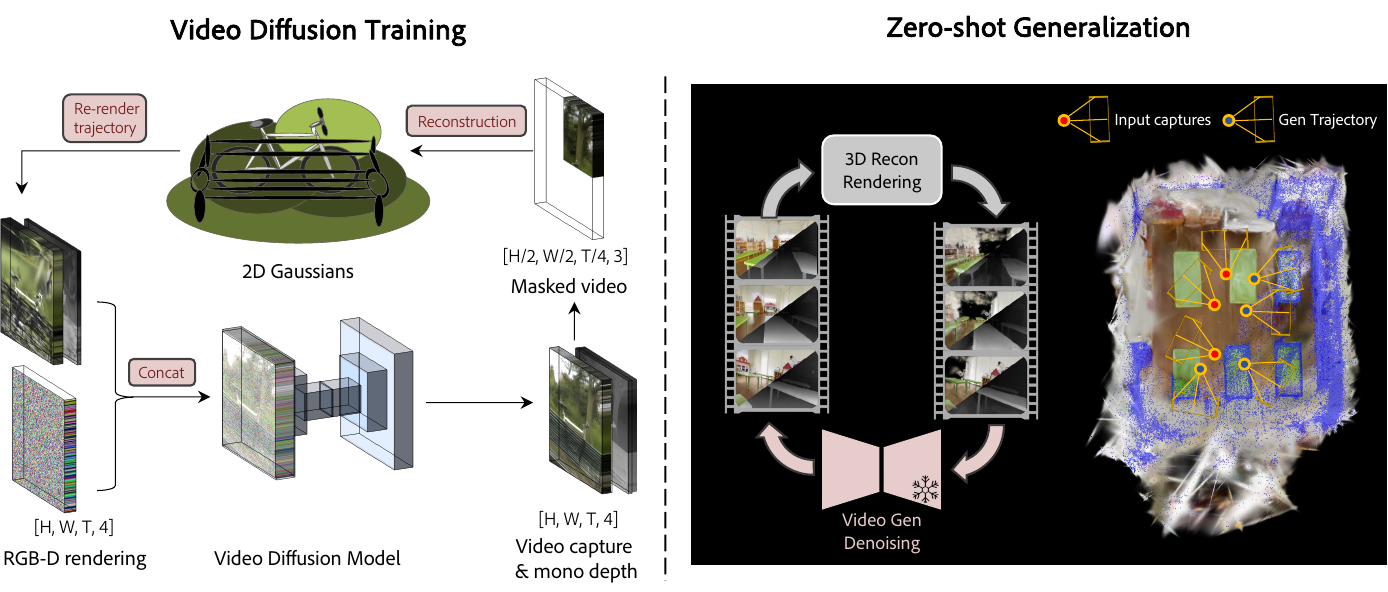} 
    \vspace{-0.1in}
    \caption{\textbf{GenFusion pipeline.} Our approach contains two stages: video diffusion pre-training (left) and zero-shot generalization (right). In pre-training, we first fine-tune DynamiCrafter~\cite{dynamicrafter} on RGB-D videos from a large-scale real-world scene video dataset~\cite{DL3DV}. Captured videos are patchified, and a random patch sequence is selected for 3D scene reconstruction, rendering full-frame RGB-D videos as input to our video diffusion model, supervised by the original video capture and its monocular depth.
    During generalization, we treat reconstruction and generation as a cyclical process, iteratively adding restoration frames from the generative model to the training set for artifact removal and scene completion.
    }
    \label{fig:pipeline}
    \vspace{-0.1in}
\end{figure*}

We introduce \textit{GenFusion}, a novel reconstruction method that leverages video generative model to achieve artifact-free 3D scene reconstruction and content expansion along novel trajectories by leveraging the proposed reconstruction pipeline, as illustrated in \figref{fig:pipeline}. The core of our approach is a simple and scalable reconstruction-driven video generation architecture that predicts realistic video from artifact-prone renderings. 
Specifically, we first fine-tune DynamiCrafter~\cite{dynamicrafter} using RGB-D videos reconstructed from a large-scale, real-world scene-level video dataset~\cite{DL3DV}. We patchify the capture videos into patches, then randomly select a patch sequence to perform 3D scene reconstruction, rendering full-frame RGB-D videos as input to our video diffusion model, which is subsequently supervised by the original video capture and its monocular depth. 
Our key insight is that masked 3D reconstruction enables flexible pre-training of video models.
As shown in \figref{fig:masked_recon}, masking $75\%$  of the input pixels during 3D reconstruction produces artifacts and missing regions sharing similar artifact patterns with far-field viewpoint rendering. The resulting artifact-prone video is encoded into latent space, with diffusion guided by a scene description token processed by a CLIP model on a randomly sampled frame.
In addition, the baseline video diffusion model, conditioned only on text and RGB, lacks sequence input handling and geometry constraints for view consistency. To address this, we embed input views as latent sequences and incorporate depth by replacing the RGB VAE with an RGB-D VAE.
Once trained, we introduce a cyclic reconstruction-generation fusion scheme for scalable 3D scene generation with artifact correction.

Our \textit{GenFusion} learns very high-capacity models that generalize well, we make the following contributions:
\begin{itemize}
\item We introduce a reconstruction-driven video diffusion model that efficiently repairs reconstruction artifacts and generates new content in invisible regions.
\item We design a masked 3D reconstruction for artifact-GT video pair generation, serving as a new novel view synthesis evaluation protocol for far-field viewpoints.
\item Experiments on challenging benchmark datasets~\cite{DL3DV,Knapitsch2017,Barron2022CVPR} demonstrate the effectiveness of GenFusion in synthesizing views distant from the training views.
\item Our approach to bridging 3D reconstruction and generation through videos is principled and straightforward, picking the best of these two fields.
\end{itemize}

\section{Related Work}
\boldparagraph{Regularization Techniques} 
Optimizing 3D scene representations, \eg, NeRF~\cite{Mildenhall2020ECCV} and 3DGS~\cite{Kerbl2023TOG}, from 2D images inherently involves an ill-posed inverse problem, where multiple solutions may exist for a given set of observations. To address these challenges, various techniques have been proposed to constrain the optimization of scene representations.
In general, unsupervised regularization techniques are based on the assumption that the 3D representation should be sparse~\cite{Hedman2021ICCV, Yu2021ICCV}, smooth~\cite{Niemeyer2022CVPR, Yang2023CVPR, Zhang2024CVPR}, low rank~\cite{Chen2022ECCV,Chen2023ARXIV,Chen2023TOG}; rendering weights to be compact~\cite{Barron2022CVPR}; or geometry/texture to be consistent with nearby views in image space~\cite{Chen2022StructNeRFNR,Fran2022CVPR, Fu2022GeoNeus}.
In addition to the multitude of regularization strategies available, many optimization techniques have been proposed to enhance training procedures.
For example, gradient scaling~\cite{gradientscaling} is applied for NeRF to address the issue of high-magnitude gradients in regions close to the camera, where more points are sampled, often leading to artifacts known as floaters. One of the most impactful approaches is the ``coarse-to-fine" training~\cite{lin2021barf,li2023neuralangelo}, which modulates the frequency band of positional encodings or hash grid resolution.
Another line of research leverages pseudo-observations by using off-the-shelf models to predict cues that enhance reconstruction quality, such as sparse point clouds~\cite{Deng2022CVPR} and monocular depth/normal maps~\cite{Yu2022NEURIPS}. These approaches are also employed in 3DGS representations~\cite{Zehao2024ECCV,Jiahe2024CVPR}. Recently, combining generation models to generate novel views as regularization has proved to be effective in reconstruction objects~\cite{Yang2024Trans} and scene~\cite{Rundi2024CVPR, liu20243dgs}. Our work aligns closely with ReconFusion~\cite{Rundi2024CVPR} and the concurrent 3DGS-Enhancer~\cite{liu20243dgs}, both of which utilize generative priors to guide the optimization of 3D representations. 
Despite these methods achieve impressive view interpolation results in sparse-view scenarios, they still struggle with rendering trajectories that deviate significantly from the input views.

\boldparagraph{Feed-forward 3D Reconstruction}
In contrast to the per-scene optimization required by neural fields, recent research has explored feed-forward architectures capable of regressing 3D scene representations from a sparse set of input images. These methods learn 3D representations from input images and directly predict novel views in a feed-forward manner.
PixelNeRF~\cite{Yu2021CVPR} predicts neural radiance representations from input images, using a convolutional network for efficient feature extraction. MVSNeRF~\cite{Chen2021ICCVb} pioneers the paradigm of leveraging cost volumes to regress realistic images from novel viewpoints. Subsequent works~\cite{Mohammad2022CVPR, Haofei2024CVPR, Yuedong2023Arxiv, Haotong2022SIGGRAPH} further improve performance through enhanced feature matching architectures. In the context of 3DGS, pixelSplat~\cite{David2024CVPR} directly regresses scene-level 3D representations from paired images, incorporating an epipolar transformer module to effectively capture view-dependent geometric correspondences. MVSplat~\cite{Yuedong2024MVSplat} builds a cost volume representation to learn cross-view feature similarity, achieving high-quality scene generation with improved efficiency. Recently,  DepthSplat~\cite{xu2024depthsplat} further boosts the performance using a pre-train multi-view depth estimator, and Long-LRM~\cite{ziwen2024llrm} utilizes Mamba2 blocks to handle many input views.
To handle single view inputs, Flash3D~\cite{Stanislaw2024Flash3D} extend monocular depth estimation to 3D shape and appearance reconstruction. Recently, SplatFormer~\cite{yutong2024splatformer} proposes a learning-based model to refine Gaussian splats, enabling out-of-distribution novel-view synthesis.
However, existing feed-forward methods are limited to a small number of views - typically fewer than $10$ - which significantly restricts their application.




\boldparagraph{View-Conditioned Generation} 
Generative models have emerged as a promising solution to synthesize plausible content for regions without observations.
Thanks to their success in image generation, diffusion models are widely used for multi-view synthesis. Early work 3DiM~\cite{Daniel2023ICLR} trains a pose-conditional image-to-image diffusion model for object-centric novel view synthesis. Follow-up research Zero-1-to-3~\cite{WeiboAAAI0123} advanced this approach by fine-tuning large-scale pre-trained diffusion models on synthetic datasets. SSDNeRF~\cite{Hansheng2023ICCV} jointly optimize diffusion and NeRF auto-decoder to synthesize novel views for object.
Recently, ZeroNVS~\cite{Kyle2024CVPR} and CAT3D~\cite{Ruiqi2024CAT3D} extended the paradigm by training diffusion models with camera pose and multi-view image conditioning.
Nevertheless, relying solely on 2D features proves insufficient for maintaining 3D consistency across generated views. Several works~\cite{Jianfeng2023, Gabriela2023} inject 3D information into diffusion models for improved geometric understanding. For instance, ViewCrafter~\cite{Wangbo2024ViewCrafter} explicitly utilizes 3D information from point clouds and expands it iteratively, facilitating consistent interpolation between two views. To enable generating long sequences of 3D scenes from a single input image, InfiniteNature-Zero~\cite{li2022_infinite_nature_zero} and following work~\cite{cai2022diffdreamer,yu2024wonderjourney,wang2024vistadream} explore single-view scene extrapolation by progressively in/outpainting layered depth images predicted by a monocular estimator.
While these generative methods achieve visually appealing 3D asset generation, their reconstruction quality and view coverage, particularly at scene scale, remains far from that of 3D scenes reconstructed from densely captured multi-view data.

\section{GenFusion}
Our goal is to construct artifact-free 3D scenes with content augmentation given conditioning views $\{\bI_i\} \, (i>=1)$. 
%
The core idea to align 3D reconstruction and generation through video renderings. Specifically, we propose a cyclic fusion approach where 3D reconstruction and generation benefit each other in a virtuous cycle: during the reconstruction process, we iteratively leverage information from the generative model to improve reconstruction quality; meanwhile, more accurate reconstruction results help the generative model produce more realistic and consistent content. This bidirectional enhancement mechanism creates a positive feedback loop.



In the pre-training stage, we aim to train a model that learns from large-scale scenes, which then serves as guidance for optimizing general ill-posed 3D lifting tasks, such as reconstructing 3D from sparse views. We use a video diffusion model as guidance, pre-trained on video captures and their artifact-prone renderings of 3D reconstructions. 
Specifically, we train a generative model $G_\phi$ by maximizing the expected log-likelihood of generating the complete image given the reconstruction from a masked input:

\begin{equation}
\arg \max_{\phi} \mathbb{E} \left[ \log p_{\phi} \left(I \mid R_\theta( \tilde{I}) \right) \right]
\label{eq:gen_objective}
\end{equation}

where $R_\theta$ denotes a reconstruction and rendering function (e.g., NeRF or 3DGS) and $\tilde{I}$ denotes the masked version of target image $I$.
To make the generation process 3D aware, our video diffusion model learns to capture the underlying distribution of images by conditioning on artifact-prone rendering video ${R}_\theta(\tilde{I})$ in our case.

With the video diffusion model, we propose optimizing the 3D scene representation, $\theta$, guided by both input image conditions and generated video for a new scene.
Specifically, we maximize the similarity between the novel view videos rendered from 3D representation and the videos generated from our pre-train diffusion model, therefore the gradient of 3D representation is propagated from the photometric loss:


\begin{equation}
\arg \min_{\theta} \mathbb{E}\left[ \left| G\phi(\hat{I}_{k+1} \mid R_\theta(\tilde{I}_{k})) - R_\theta(\hat{I}_{k+1}) \right|_2 \right]
\label{eq:recon_objective}
\end{equation}
where $k$ denotes the iteration index.

In the following sections, we first introduce our reconstruction-driven generation architecture, \ie ${G}_{\phi}(I|R_{\theta}(\tilde{I}))$, in Sec.~\ref{sec:generation}. Next, we detail our cyclic generation and reconstruction fusion process in Sec.~\ref{sec:fusion}.

\subsection{Reconstruction-driven Generation}
\label{sec:generation}
We generate view-consistent video by conditioning the current fragment's generation on previous corrupted reconstructions. To generate corrupted reconstructions, we propose a masked 3D reconstruction approach for generating training data, along with a new video diffusion model that facilitates 3D-aware generation and regularization.

\begin{figure}
    \centering
    \includegraphics[width=0.5\textwidth]{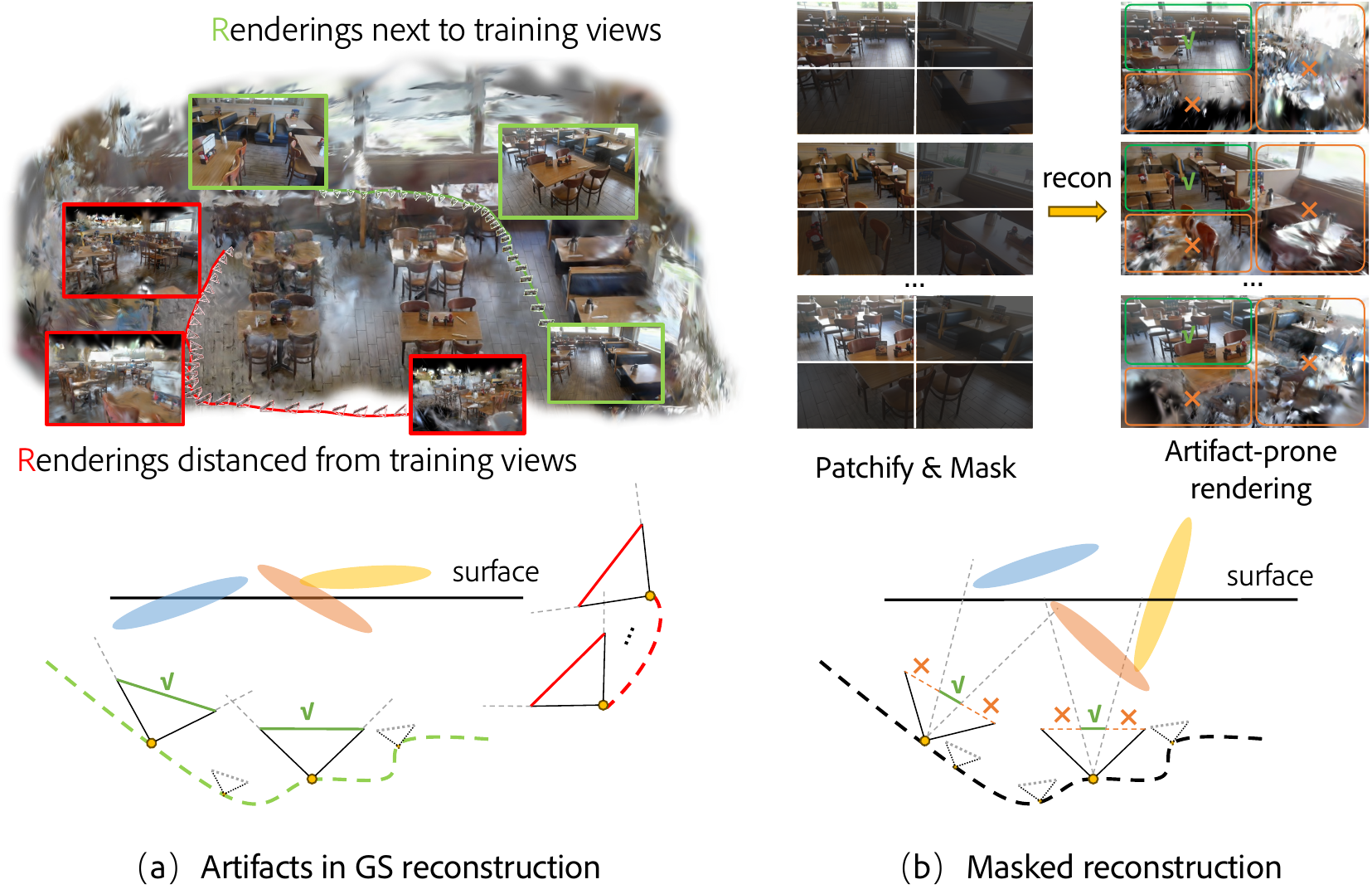} 
    \caption{\textbf{Artifact-GT video pair generation using masked reconstruction.} a) current SOTA Gaussian Splatting methods render accurately near training views but produce artifacts for distant views due to limited angular supervision, like the red trajectory. b) we propose a masked reconstruction scheme to replicate such artifact patterns for training video diffusion models by masking $75\%$ of pixels during 3D reconstruction and re-rendering the scene along the original trajectory, including the masked pixels.}
    \label{fig:masked_recon}
    \vspace{-0.1in}
\end{figure}

\begin{figure*}[t!]
    \centering
    \includegraphics[width=0.98\textwidth]{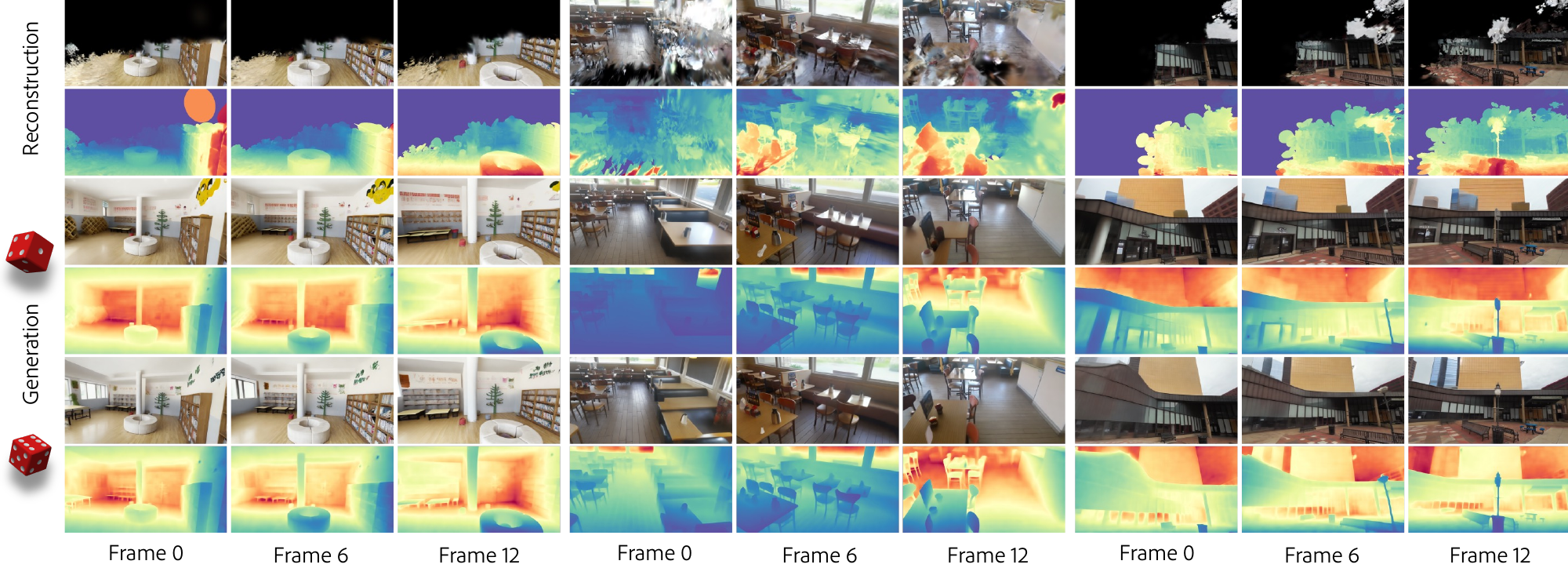} 
    \vspace{-0.1in}
    \caption{\textbf{Reconstruction-driven Video Generation.} Our video diffusion model is able to generate realistic RGB-D video from artifact-prone RGB-D renderings, which is then used as photometric guidance in our cyclic fusion period.}
    \label{fig:video_generation}
\end{figure*}

\boldstartspace{Masked 3D Reconstruction.}
We now discuss how to learn novel view generalization capability in a reconstruction paradigm from open-world large-scale videos.
Given video captures of a scene, a straightforward approach is to downsample frames at uniform intervals or split the input sequence at the midpoint into train/test segments for scene reconstruction, then render the reconstructed scene from the testing camera viewpoints to generate artifact-prone data. These rendered sequences and original video captures then serve as input and output pairs for video diffusion model training.
However, we observe that these sampling schemes limit models to either view interpolation or pure generation. Specifically, scenes are often fully covered by sampled views, with target views located adjacent to these, while the alternative approach often leaves most content unseen in the training segment.
Our work aims to learn regularization and generate new content in trajectories that deviate significantly far from the capture trajectories.

To this end, we propose a masked 3D reconstruction technique to obtain corrupted 3D scenes from observations across both spatial and temporal dimensions, then render images for the full input sequence to generate the training dataset.
In specific, we divide input image captures into $4$ regular non-overlapping patches, \ie top-left, top-right, bottom-left and bottom-right, as shown in \figref{fig:masked_recon}. Then we sample the top-left or bottom-right patch region and mask out (i.e., remove) the remaining three for each scene. The sampled patch sequence is used to conduct 3D reconstruction using the standard approaches, \ie 2D Gaussian Splatting (2DGS)~\cite{Huang2DGS2024} in this paper.
Note that the mask is applied per scene rather than per view, forcing the reconstruction process to have limited view coverage. In practice, since we use open-world large-scale video sequence captures $\bI$ for training, which further increases sparsity $-$ views are dense only on the trajectory but extremely sparse in angle coverage $-$ unlike standard multi-view datasets such as Mip-NeRF 360~\cite{Barron2022CVPR}. 

With masked reconstruction, we render full-view videos along the same camera poses as the input to form real capture and reconstruction rendering pairs for our video diffusion model training.
Our insight is that reconstructing from masked images simulates a narrower field-of-view camera, requiring that context outside the mask be inferred from views deviating from the current viewpoint, which promotes view extrapolation.
Moreover, rendering to full view introduces unconstrained regions with extensive black backgrounds, facilitating content outpainting.

\boldstartspace{Video Diffusion.}

In essence, we build our video generation model upon the foundation of pre-trained DynamiCrafter~\cite{dynamicrafter} - an image-to-animation model - and adapt the model to reconstruction-related video restoration tasks.

More specifically, we enhance frame consistency by incorporating geometric information into the generation process. This is achieved by replacing the VAE in the baseline model with a pre-trained RGB-D VAE, where encoder and decoded are denoted as $\mathcal{E}$ and $\mathcal{D}$ respectively. It allows depth to be integrated without altering the diffusion architecture.
%
In the training process, the ground truth RGBD video $I_{RGB-D}$ are encoded into latent space $z := \mathcal{E}(I_{RGB-D})$, which we add noise in different timestep $t$ and obtain $z_t$. To guide the generation process under reconstruction result, We provide two conditions $c$. The artifact-prone RGB-D video ${I}^{'}_{RGB-D}$ is encoded and concatenated with per-frame initial noise to support sequence conditioning.
This allows rich visual details from the rendered videos to condition the generation process effectively.
Additionally, we select the input view closest to the rendering trajectory and embed its ground-truth image into the CLIP feature space. This high-level conditioning provides global information about the scene content.
Thus the video denoising network $\epsilon_\theta$ is optimized by:

\begin{equation}
\mc{L} = \mathbb{E}_{\mathcal{E}(x),c,\epsilon\sim\mathcal{N}(0,1),t}\left[\left\|\epsilon - \epsilon_\theta(z_t, t, c)\right\|_2^2\right],
\end{equation}
The refined latent is subsequently decoded into the final RGB-D video through a pre-trained VAE decoder $\hat{V} =\mathcal{D}(z)$~\cite{Gabriela2023}.

\subsection{Cyclic Fusion}
\label{sec:fusion}

We build our fusion process upon the popular 2D Gaussian Splatting and use a number of 2D oriented planar Gaussian disks to represent 3D scenes. Gaussian disks are parameterized by center position $p \in \mathbb{R}^{3\times 1}$, opacity (scale) $\alpha \in [0,1]$, two principal tangential vectors $\bt_u$ and $\bt_v$ for orientation, a calling vector $\bS=(s_u,s_v)$, and Spherical Harmonices (SH) coefficients. We refer the reader to the original paper for representation and splatting details~\cite{Huang2DGS2024}. 
In the following, we introduce how to initialize and progressively update the Gaussian primitives by fusing the 3D reconstruction and video diffusion output.



\boldstartspace{Fusion.} We follow the original 3DGS approach by reusing the calibration point cloud for initialization. We initialize Gaussians and their attributes, which are then updated end-to-end based on rendering losses. The optimization process operates as a reconstruction and generation cycle, supervising Gaussians with both input conditioning and novel generated views. More concretely, for every $K$ iteration, we begin with sample new trajectories and render RGB-D videos based on the current reconstruction. We feed these rendering sequences to our video diffusion model to generate artifact-free videos, which are then added to the supervision set, as shown in~\figref{fig:pipeline}. The cyclic process enables artifact correction in under-observed regions and generates new content for areas that are invisible within the input views.

Among these, we find that novel trajectory sampling is the most critical component. To ensure comprehensive view and angle coverage, we employ two types of trajectories: view interpolation between neighboring input views and a spiral/spherical path generated across all input camera poses.

\boldstartspace{Content Expansion.} 
Large unobserved areas can appear as black or noisy pixels when the sampling trajectories are away from the input views. Although we use generation outputs as supervision for these regions, we found it challenging to split and clone new Gaussians~\cite{Kerbl2023TOG} due to the absence of surrounding Gaussians. 
We solve this issue by adaptively adding new Gaussian points to the scene during optimization, using an unreliable depth-based mapping. Pixels are considered unreliable when

\begin{align}
T < \tau_T \text{ or } |D - \hat{D}| > \tau_D
\end{align}
where $T$ is cumulative opacity, $D$ and $\hat{D}$ are rendered depth and aligned generated depth D respectively. $\tau_T$ and $\tau_D $ are hyperparameter thresholds.

For these unreliable areas, we add new Gaussians by back-projecting the generated RGB-D points into 3D space, similar to the initialization stage. Note that for the newly added Gaussians, position and color values are directly obtained from the RGB-D video, while other attributes are initialized as in 2DGS. 


\boldstartspace{Loss function.}
During cyclic fusion, we freeze the video diffusion model and optimize the 3D representation end-to-end, using simple photometric losses between rendered RGB-D images and input (\ie $\mathcal{L}_{recon}$) and generated views (\ie $\mathcal{L}_{gen}$):
 
\begin{align}
\mathcal{L} = \mathcal{L}_{recon}  + \lambda \mathcal{L}_{gen}
\end{align}
where $\mathcal{L}_{recon} = \lambda{_{l_1}} \cdot \mathcal{L}_{l1} + \lambda{_{SSIM}} \cdot \mathcal{L}_{SSIM} + \lambda{_{mono}} \cdot \mathcal{L}_{mono}$, and the $\mathcal{L}_{mono}$ is a scale-invariant depth loss, as used in~\cite{Yu2022NEURIPS} that enforces consistency between our
rendered rendered depth $\hat{D}$ and the monocular depth  $D$ predicted by our video diffusion model. The generation loss $\mathcal{L}_{gen}$ shares components but is instead applied to generated images.

To stabilize the optimization process, we use an Sinusoidal warm-up and annealing strategy for the generation loss weight $\lambda$, which is defined as:

\begin{align}
\lambda(k) = 1.0 \cdot \sin\left( \frac{k - K_{\text{start}}}{K_{\text{end}} - K_{\text{start}}} \cdot \pi \right)
\end{align}
where $k$ is the iteration, and $K_\text{start}$ and $K_\text{end}$ are the start and end iteration of the diffusion term.

\section{Experiments}
We begin with the experimental setup for training and evaluating our GenFusion. Next, we present our reconstruction-driven video diffusion model and compare it quantitatively/qualitatively with state-of-the-art methods for sparse-view 3D reconstruction and view synthesis. Finally, we demonstrate the generative capabilities of our method with scene completion. 

\subsection{Experimental Setup}

\boldparagraph{Training set} 
We train our diffusion model on DL3DV-10K~\cite{DL3DV}, a large-scale dataset containing 10,510 videos, including 140 benchmark scenes. To prepare reconstruction-gt video pairs for diffusion model training, we optimize each scene for 7K steps using our masked 3D reconstruction scheme introduced in \secref{sec:generation}. The number of training views are uniformly downsampled to $\nicefrac{1}{4}$ frames for each video to ensure sufficient sparsity, and we render the reconstructed scenes along the original trajectories at a resolution of $960 \times 540$ and augment the dataset with depth information using the current SOTA monocular depth estimator~\cite{hu2024-DepthCrafter}.

\boldparagraph{Evaluation Dataset} 
We evaluate our method on diverse scenes from three datasets, including 24 scenes from the DL3DV-Benchmark~\cite{DL3DV}, 7 scenes from Tanks and Temples (TnT)~\cite{Knapitsch2017}, and 9 scenes from the Mip-NeRF360~\cite{Barron2022CVPR} dataset, see supplement for details.


\boldparagraph{Implementation Details}
In this work, we adopt DynamiCrafter~\cite{dynamicrafter} as the video diffusion backbone and fine-tune the model on artifact-GT RGB-D video pairs. The fine-tuning includes both coarse and fine stages: in the coarse stage, the video resolution is set to $16{\times}320{\times}512{\times}4$ with a latent space dimension of $16{\times}40{\times}64{\times}4$. We train this stage for 30K steps with a learning rate of $1{\times}10^{-5}$ and a batch size of 2 on four H100 GPUs. The model is then fine-tuned to a higher resolution of $16{\times}512{\times}960{\times}4$ for another 34K steps in the fine stage. To handle the RGB-D format, we use a frozen RGB-D VAE from LDM3D~\cite{Gabriela2023}.
During inference, we apply DDIM sampling with 25 steps and set the classifier-free guidance scale to $3.2$.

For zero-shot generalization, we use 2DGS as the 3D representation and initialize with the COLMAP point cloud. In our experiments, views are masked and frames are downsampled; we filter and retain only the points visible from the training views for point cloud initialization.




\subsection{Video Generation}

In this paper, we introduce a reconstruction-driven video diffusion architecture that enables novel view regularization and content generation. We report the VAE design and its impact on video resolution in~\tabref{tab:fid_comparison}. Surprisingly, we find that our model fine-tuned with RGB-D VAE achieves a better FID score than the RGB VAE, even though the diffusion backbone was originally pre-trained on the RGB VAE latent space. Significant improvement can be observed when increasing the spatial resolution from $512\times320$ to $960\times512$. We show the visualization results of our diffusion model in~\figref{fig:video_generation}, our video diffusion model effectively removes ``floaters" from the input video while generating realistic content in black-pixel regions.

\begin{table}
\centering
\resizebox{1.0\linewidth}{!}{
\begin{tabular}{l|cccc}
Method & RGB VAE & RGB-D VAE & RGB-D VAE &RGB-D VAE  \\
\hline
Frames & 16 & 16 & 48 & 16\\
Resolution & 512$\times$320 & 512$\times$320 & 512$\times$320 & 960$\times$512 \\
FID$\downarrow$ & 26.1607 & 25.4006 & 29.3545 & 22.5526 \\
\end{tabular}}
\caption{\textbf{Analysis on reconstruction-driven video diffusion model.}}
\label{tab:fid_comparison}
\end{table}

\begin{table}[]
    \renewcommand{\tabcolsep}{2pt}
    \centering
    \resizebox{1.0\linewidth}{!}{
    \begin{tabular}{@{}l@{\,\,}|cccc|cccc|cccc}
    & \multicolumn{4}{c|}{PSNR $\uparrow$} & \multicolumn{4}{c|}{SSIM $\uparrow$} & \multicolumn{4}{c}{LPIPS $\downarrow$}  \\
    & 3-view & 6-view & 9-view & Avg. & 3-view & 6-view & 9-view & Avg. & 3-view & 6-view & 9-view & Avg.  \\ \hline

Zip-NeRF~\cite{Jonathan2023ICCV} & 12.77 & 13.61 & 14.30 & 13.56 & 0.271 & 0.284 & 0.312 & 0.289 & 0.705 & 0.663 & 0.633 & 0.667 \\
DiffusioNeRF~\cite{DiffusioNeRF} & 11.05 & 12.55 & 13.37 & 12.32 & 0.189 & 0.255 & 0.267 & 0.237 & 0.735 & 0.692 & 0.680 & 0.702 \\
FreeNeRF~\cite{Yang2023CVPR} & 12.87 & 13.35 & 14.59 & 13.60 & 0.260 & 0.283 & 0.319 & 0.287 & 0.715 & 0.717 & 0.695 & 0.709 \\
SimpleNeRF~\cite{somraj2023simplenerf} & 13.27 & 13.67 & 15.15 & 14.03 & 0.283 & 0.312 & 0.354 & 0.316 & 0.741 & 0.721 & 0.676 & 0.713 \\
ZeroNVS~\cite{Kyle2024CVPR} & \tbest 14.44 & 15.51 & 15.99 & 15.31 & 0.316 & 0.337 & 0.350 & 0.334 & 0.680 & 0.663 & 0.655 & 0.666 \\
ReconFusion~\cite{Rundi2024CVPR} & \best 15.50 & \sbest 16.93 & \sbest 18.19 & \sbest 16.87 & \sbest 0.358 & \tbest 0.401 & 0.432 & \tbest 0.397 & 0.585 & 0.544 & 0.511 & 0.547 \\
\hline
3DGS~\cite{Kerbl2023TOG} & 13.06 & 14.96 & 16.79 & 14.94 & 0.251 & 0.355 & \tbest 0.447 & 0.351 & \best 0.576 & \sbest 0.505 & \best 0.446 & \best 0.509 \\
2DGS~\cite{Huang2DGS2024} & 13.07 & 15.02 & 16.67 & 14.92 & 0.243 & 0.338 & 0.423 & 0.335 & \tbest 0.580 & \tbest 0.506 & \sbest 0.449 & \sbest 0.512 \\
FSGS~\cite{Zhang2024CVPR} & 14.17 & \tbest 16.12 & \tbest 17.94 & \tbest 16.08 & \tbest 0.318 & \sbest 0.415 & \sbest 0.492 & \sbest 0.408 & \sbest 0.578 & 0.517 & 0.468 & 0.521 \\
GenFusion (Ours) & \sbest 15.29 & \best 17.16 & \best 18.36 & \best 16.93 & \best 0.369 & \best 0.447 & \best 0.496 & \best 0.437 & 0.585 & \best 0.500 & \tbest 0.465 & \tbest 0.517
    \end{tabular}
    }
    \caption{\textbf{Quantitative evaluation of sparse view 3D reconstruction methods on Mip-NeRF360 dataset.} Our approach demonstrates strong performance across a variety of domains, surpassing baseline methods in most cases. We color each column as: \colorbox{tablered}{best}, \colorbox{orange}{second best}, and \colorbox{yellow}{third best}. The NeRF baseline results above are taken from Reconfusion.}
     \label{tab:sparse_view}
\end{table}

\newcommand{\multiwidth}{0.15\textwidth}

\begin{figure}[t]
    \centering
    \setlength{\tabcolsep}{0.1em}
    \hfill{}\hspace*{-0.5em}
    \renewcommand{\arraystretch}{0.4}
    \scriptsize
    \begin{tabular}{cccc}
    {} & 9 views &6 views &3 views\\
   \multirow{1}{*}[13ex]{\rotatebox{90}{3DGS~\cite{Kerbl2023TOG}}} &
    \includegraphics[width=\multiwidth]{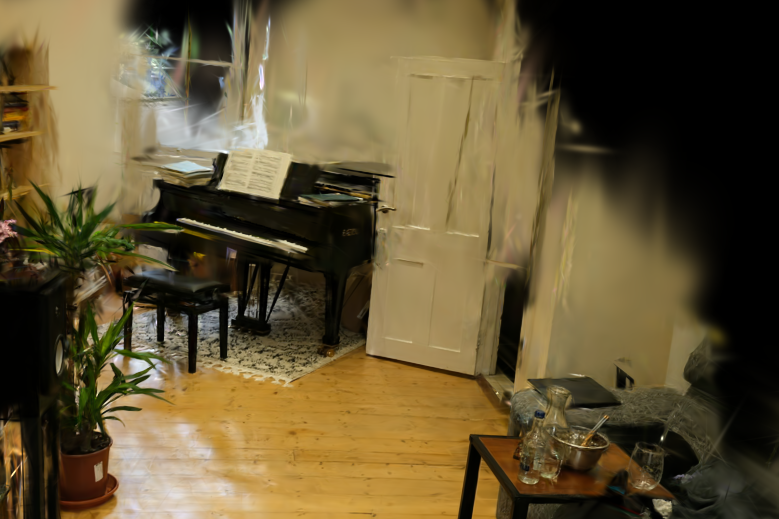} &
    \includegraphics[width=\multiwidth]{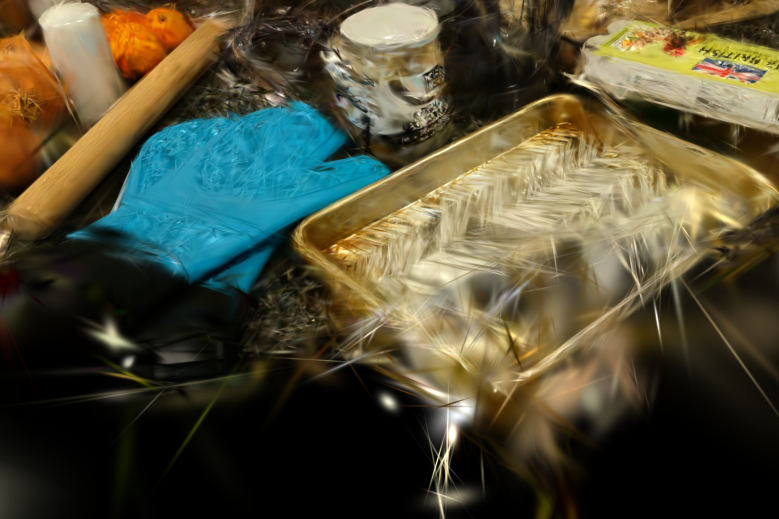} &
    \includegraphics[width=\multiwidth]{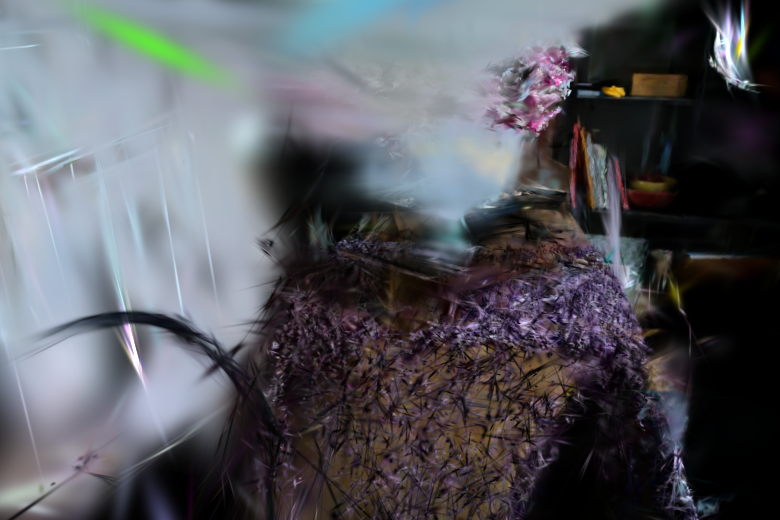} \\
    \multirow{1}{*}[13ex]{\rotatebox{90}{2DGS~\cite{Huang2DGS2024}}} &
    \includegraphics[width=\multiwidth]{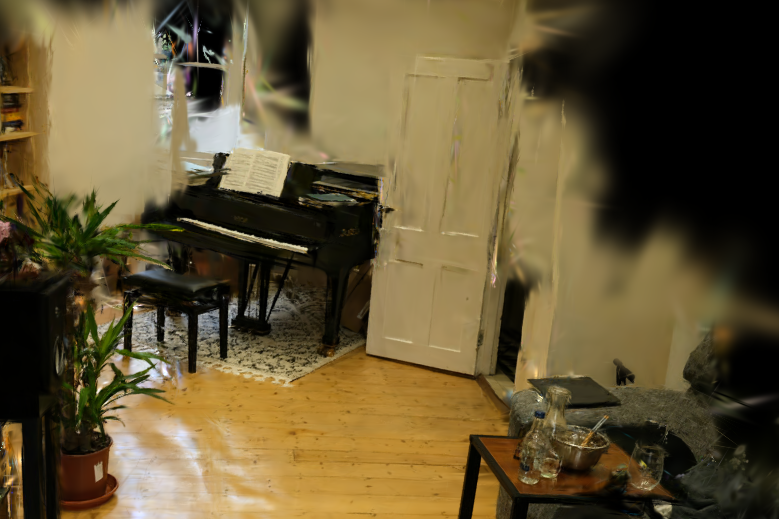} &
    \includegraphics[width=\multiwidth]{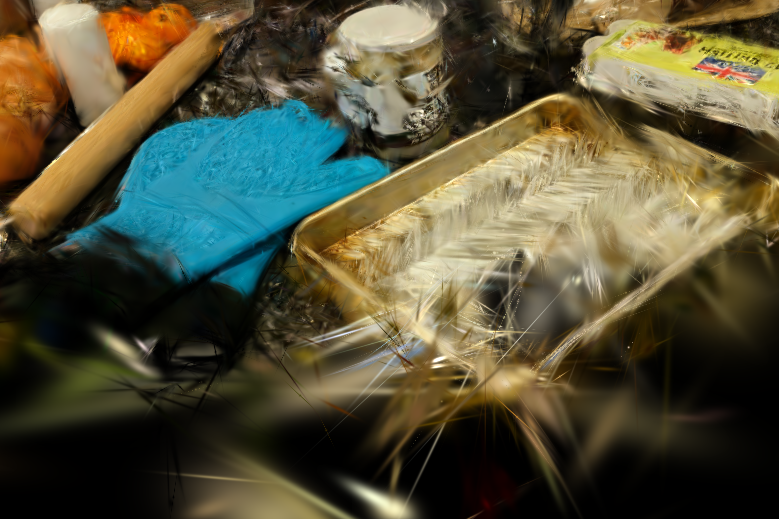} &
    \includegraphics[width=\multiwidth]{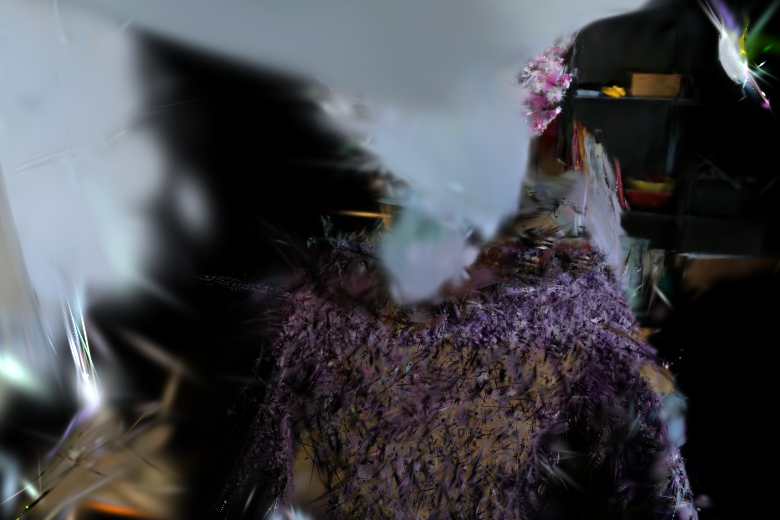} \\
    \multirow{1}{*}[13ex]{\rotatebox{90}{FSGS~\cite{Zehao2024ECCV}}} &
    \includegraphics[width=\multiwidth]{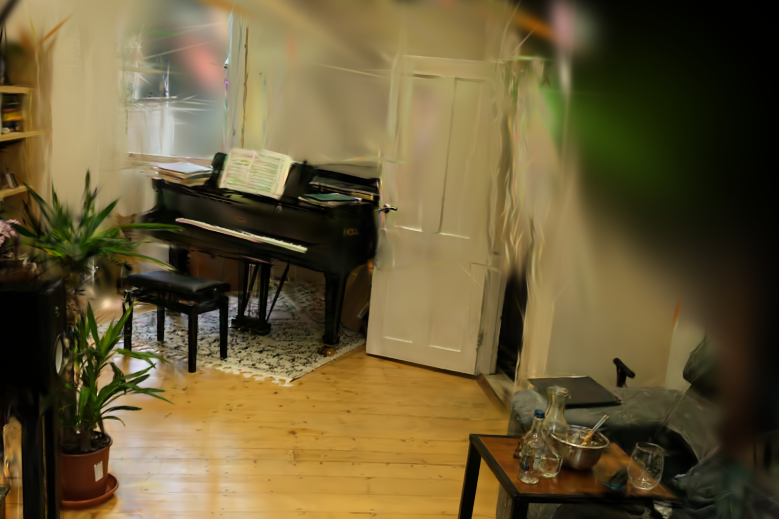} &
    \includegraphics[width=\multiwidth]{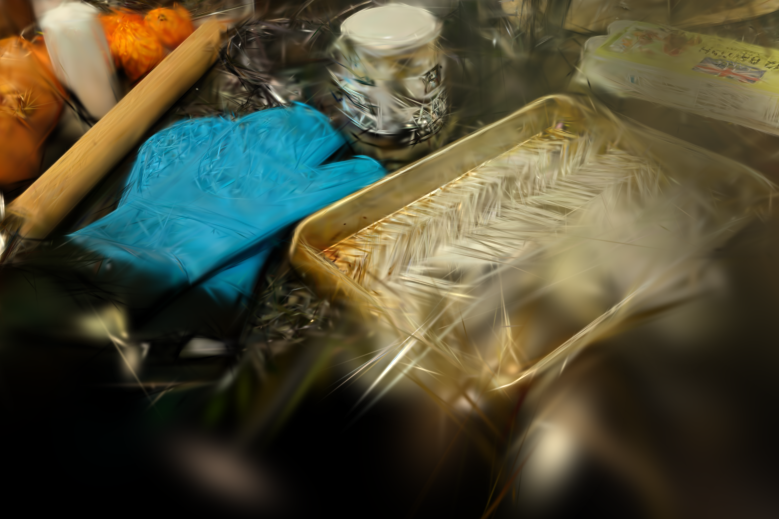} &
    \includegraphics[width=\multiwidth]{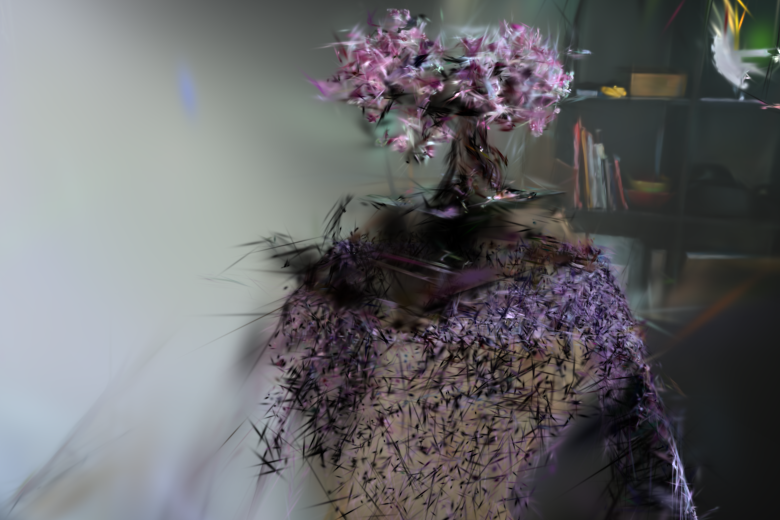} \\
    \multirow{1}{*}[9ex]{\rotatebox{90}{Ours}} &
    \includegraphics[width=\multiwidth]{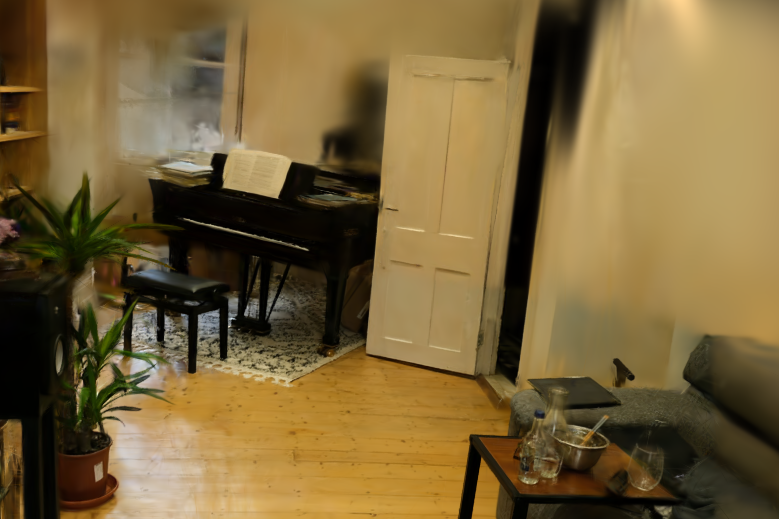} &
    \includegraphics[width=\multiwidth]{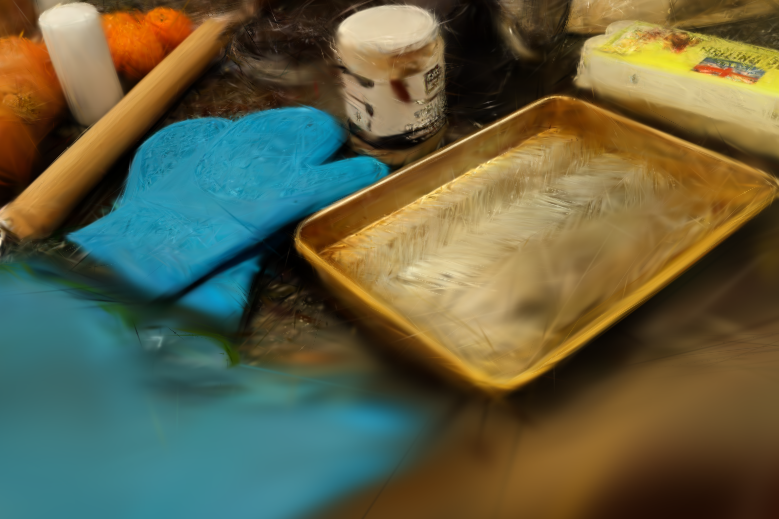} &
    \includegraphics[width=\multiwidth]{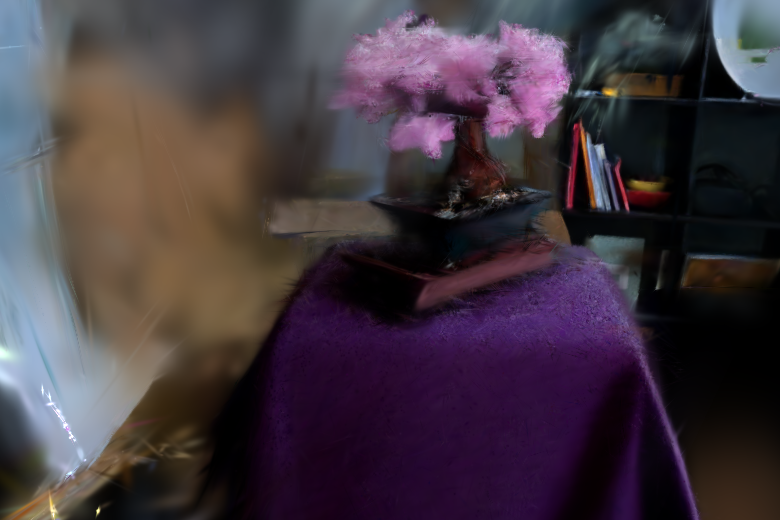} \\
    \multirow{1}{*}[9ex]{\rotatebox{90}{GT}} &
    \includegraphics[width=\multiwidth]{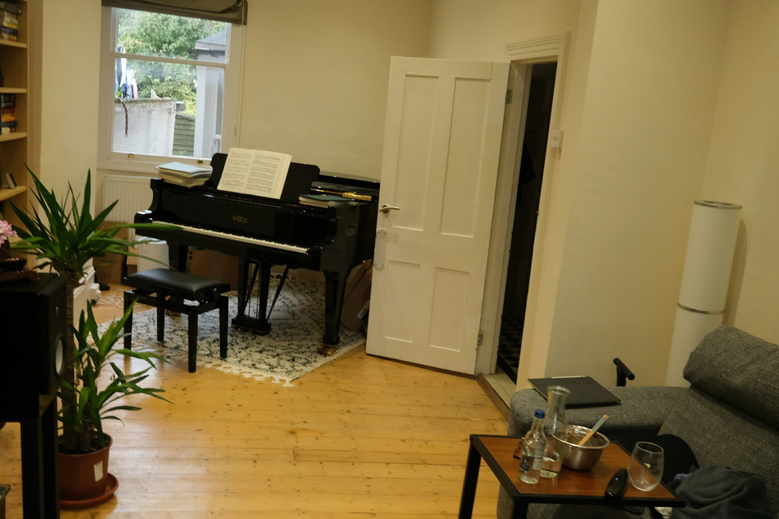} &
    \includegraphics[width=\multiwidth]{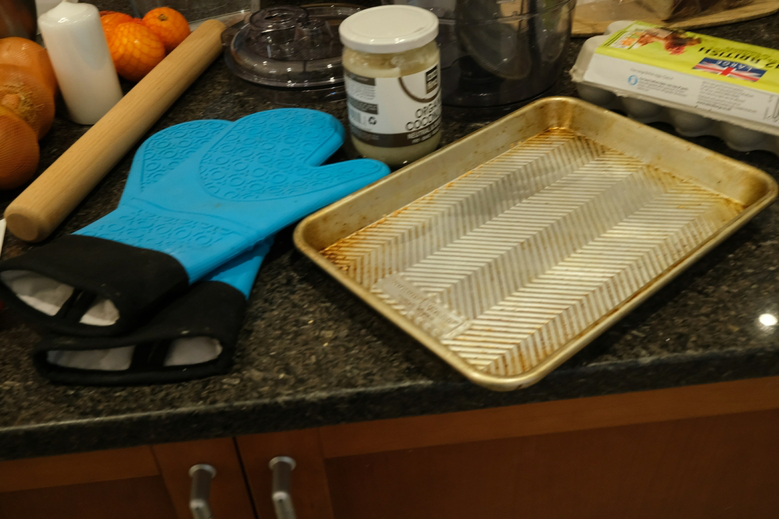} &
    \includegraphics[width=\multiwidth]{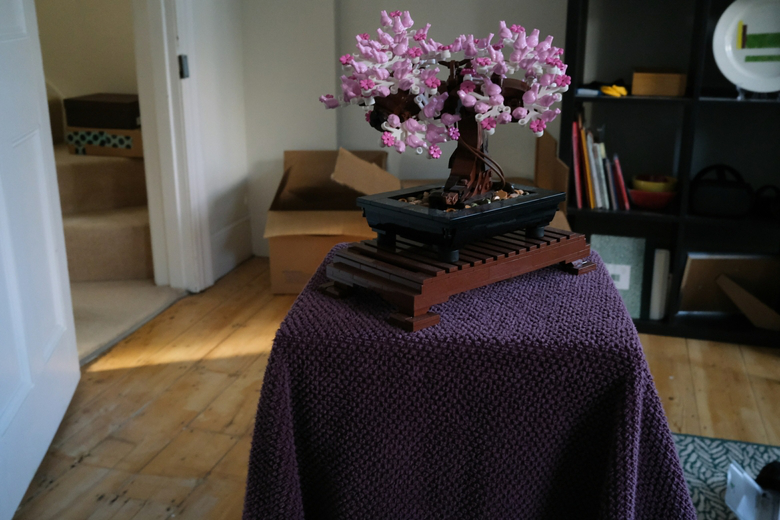} \\
\end{tabular}

    \vspace{-0.1in}
    \caption{
    \textbf{\textbf{Qualitative comparison of novel view synthesis using sparse view input on Mip-NeRF360 scenes~\cite{Barron2022CVPR}.}} 
    }
    \label{fig:sparse_view}
    \vspace{-0.1in}
\end{figure}



\begin{figure*}[htbp]
    \vspace{-0.2in}
    \centering
    \setlength{\tabcolsep}{0.2em}
      \renewcommand{\arraystretch}{0.4}
    \begin{tabular}{cccccc}
        \includegraphics[width=0.19\textwidth]{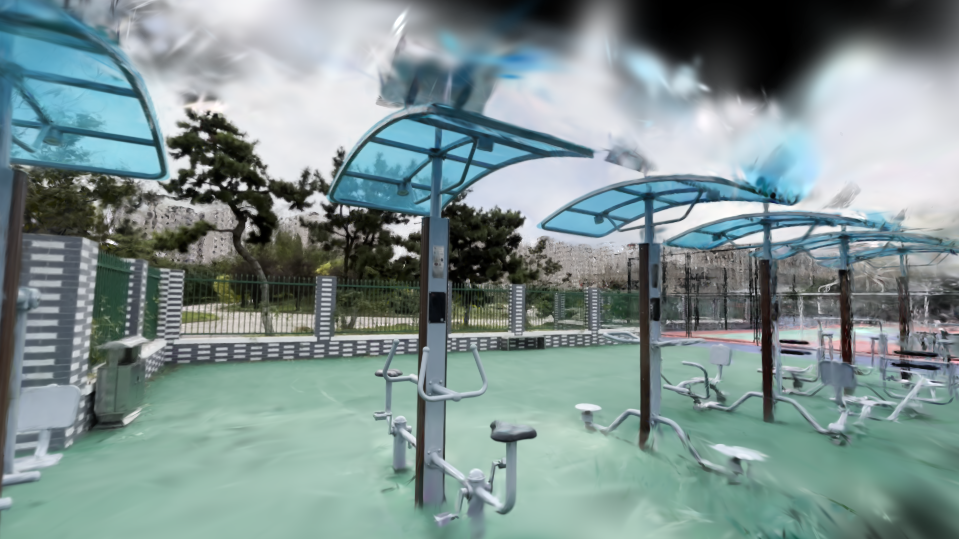} &
        \includegraphics[width=0.19\textwidth]{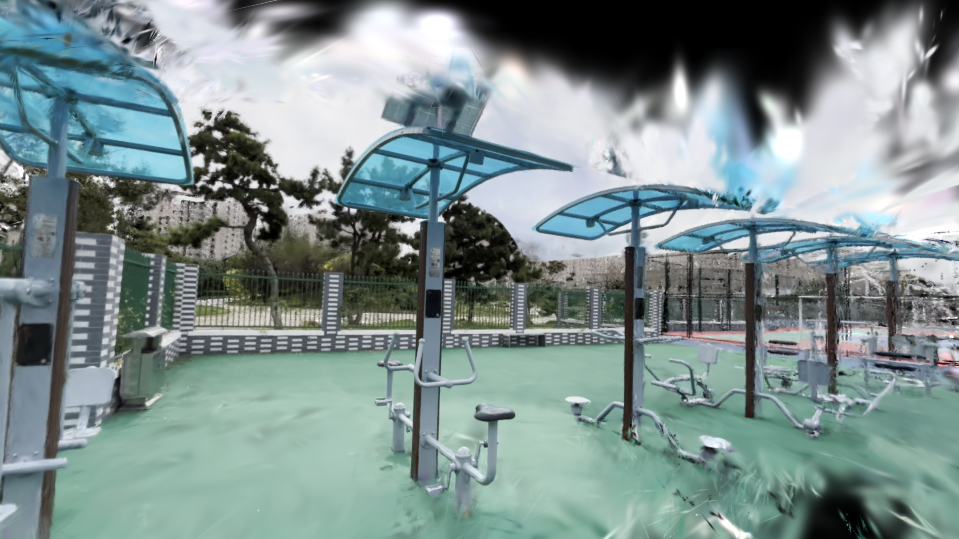} &
        \includegraphics[width=0.19\textwidth]{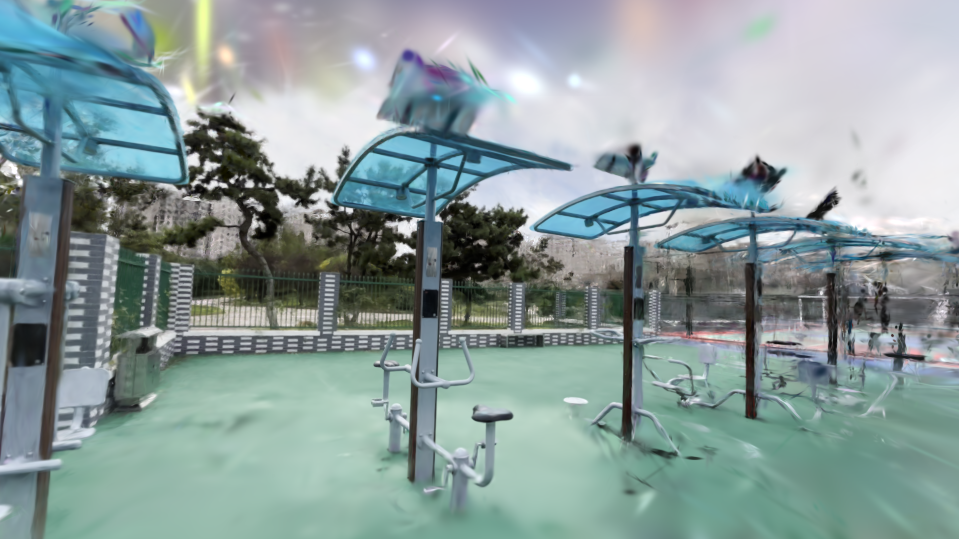} &
        \includegraphics[width=0.19\textwidth]{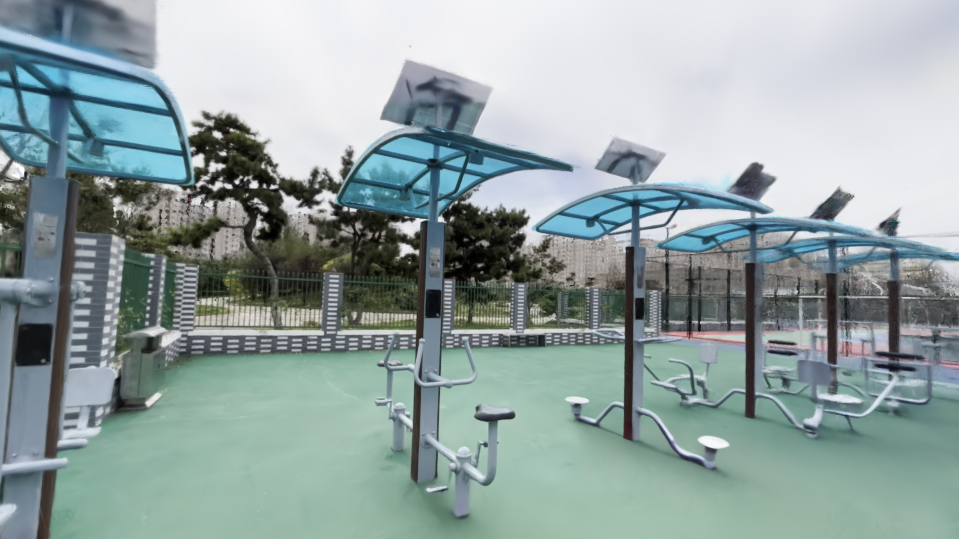} &
        \includegraphics[width=0.19\textwidth]{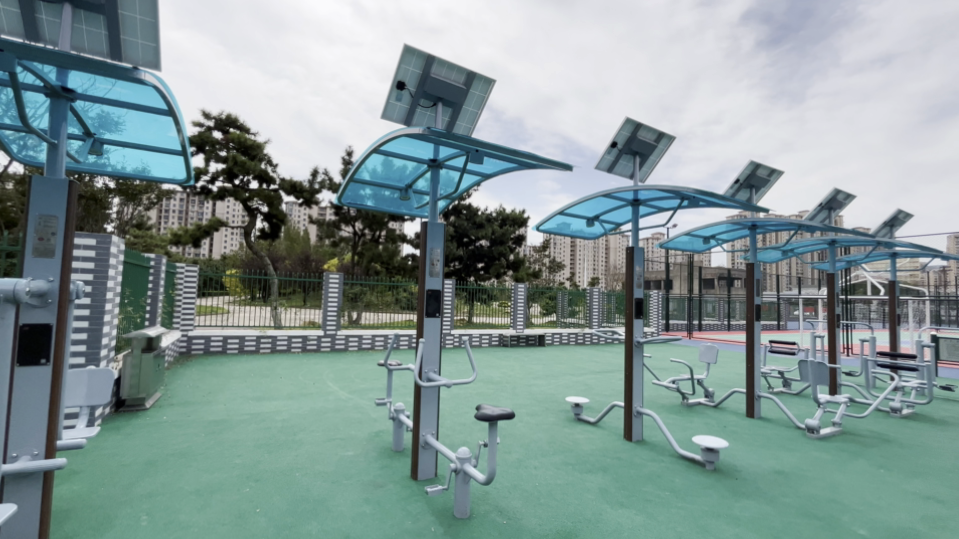} \\
        
        \includegraphics[width=0.19\textwidth]{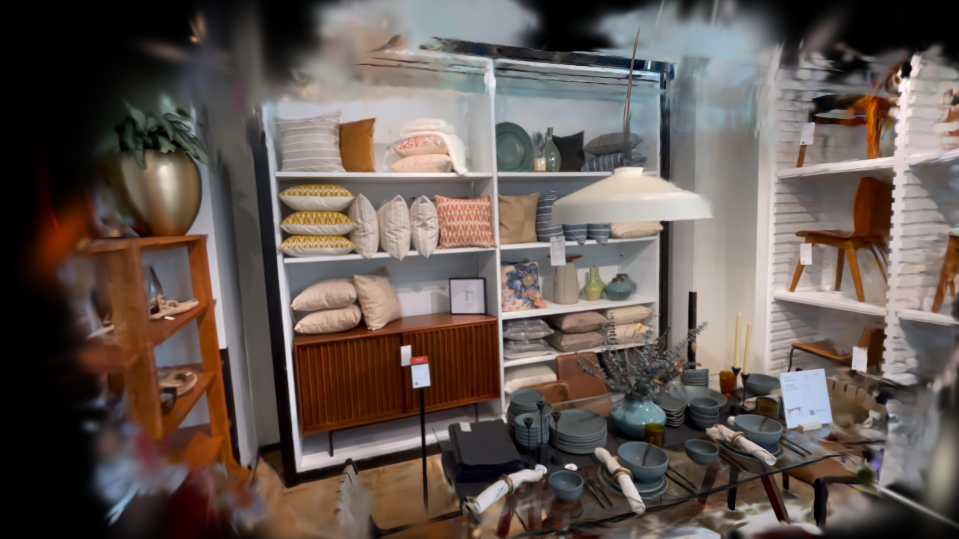} &
        \includegraphics[width=0.19\textwidth]{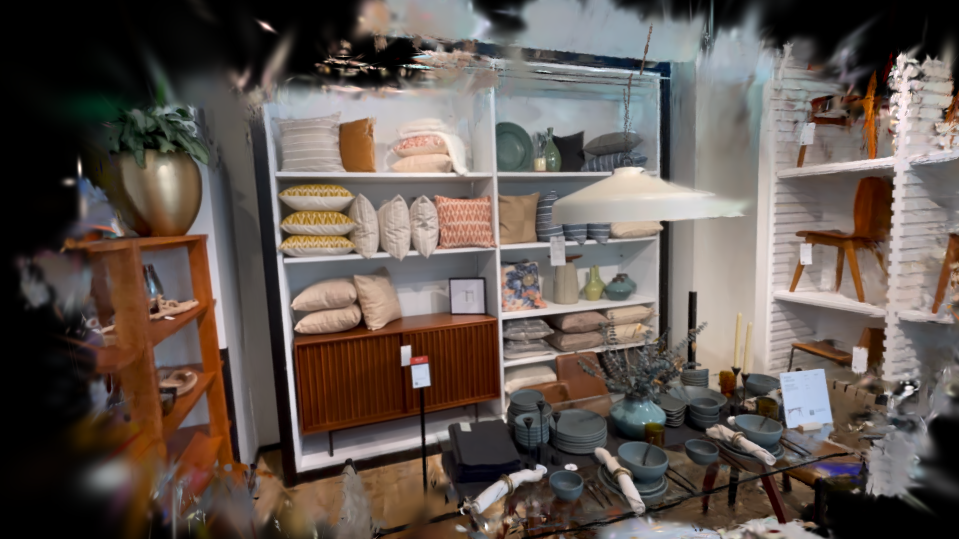} &
        \includegraphics[width=0.19\textwidth]{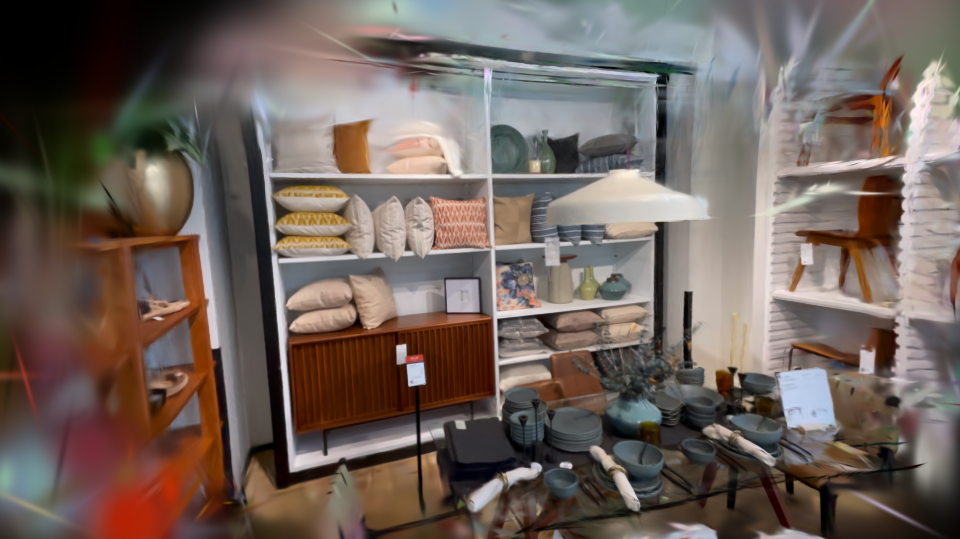} &
        \includegraphics[width=0.19\textwidth]{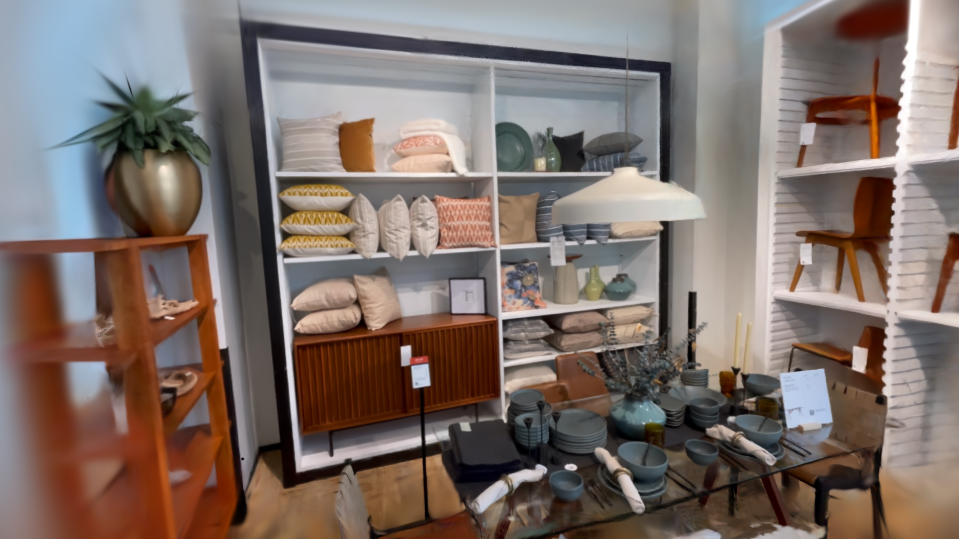} &
        \includegraphics[width=0.19\textwidth]{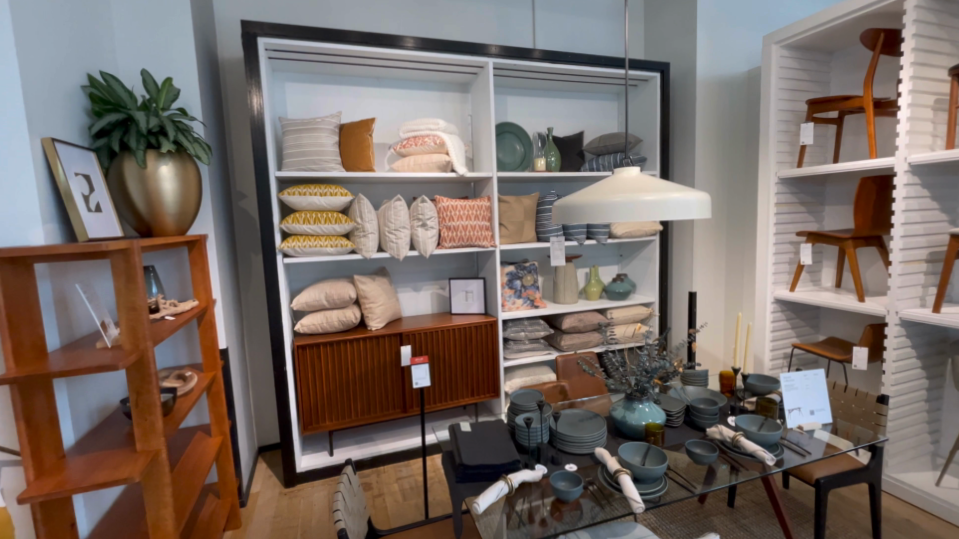} \\

        \includegraphics[width=0.19\textwidth]{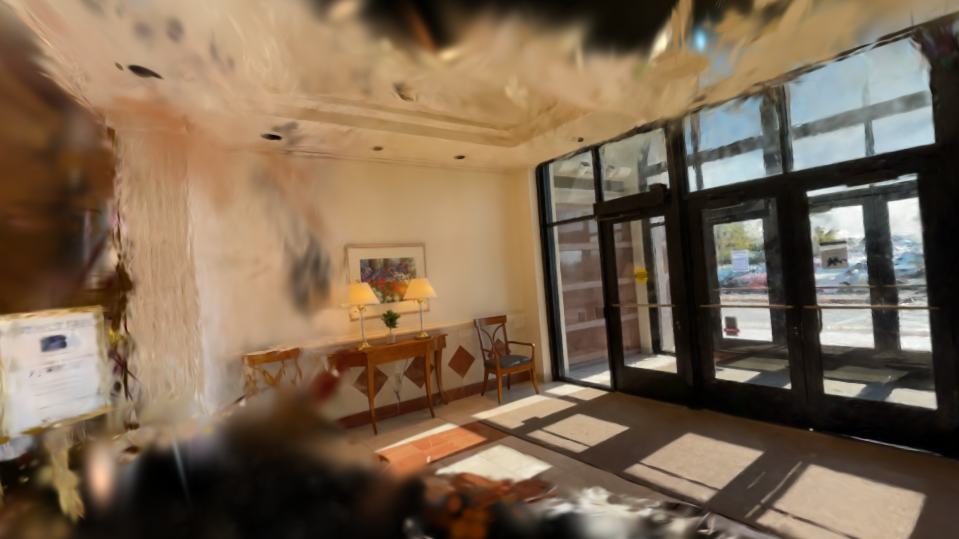} &
        \includegraphics[width=0.19\textwidth]{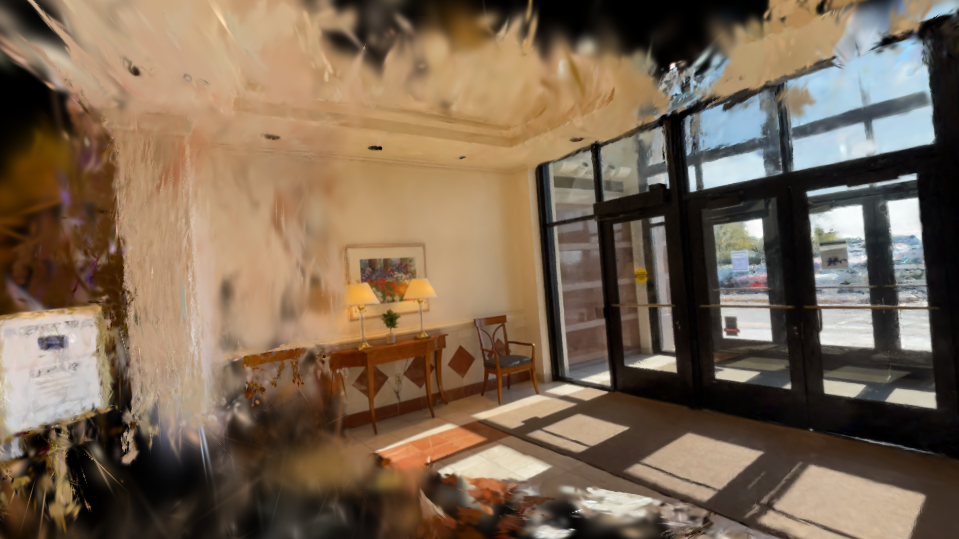} &
        \includegraphics[width=0.19\textwidth]{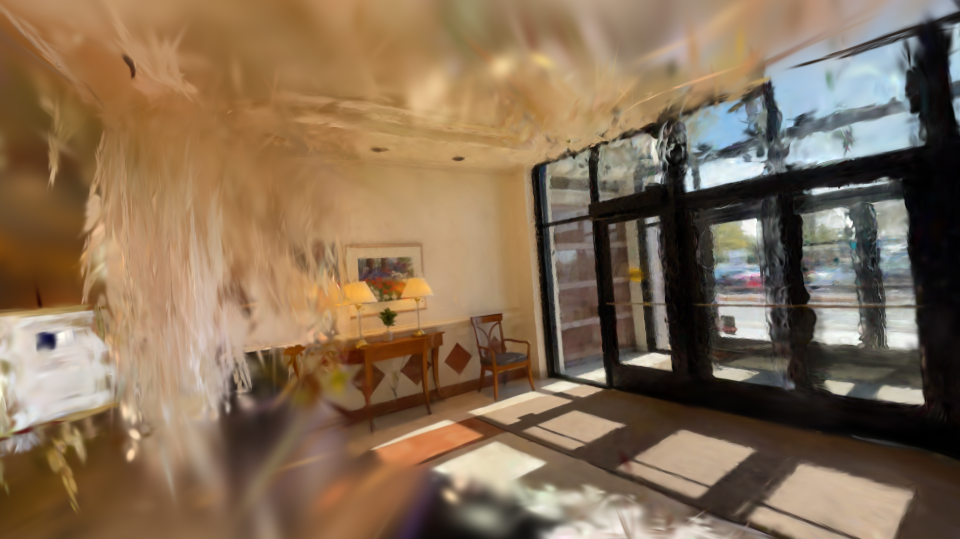} &
        \includegraphics[width=0.19\textwidth]{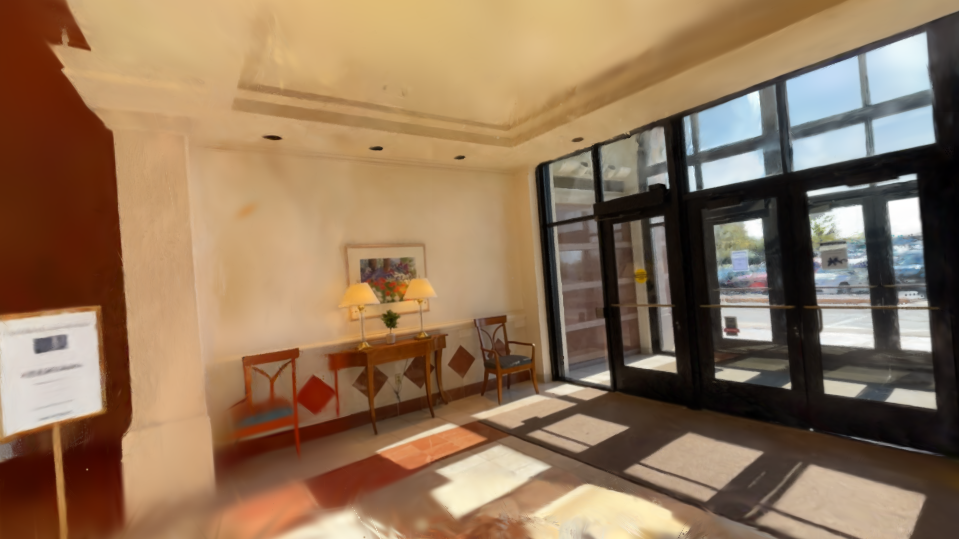} &
        \includegraphics[width=0.19\textwidth]{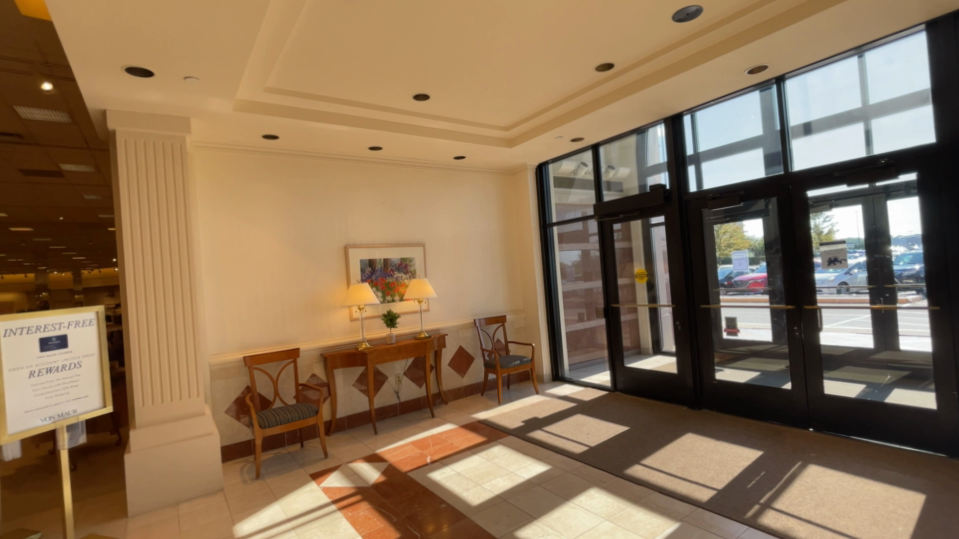} \\

        \includegraphics[width=0.19\textwidth]{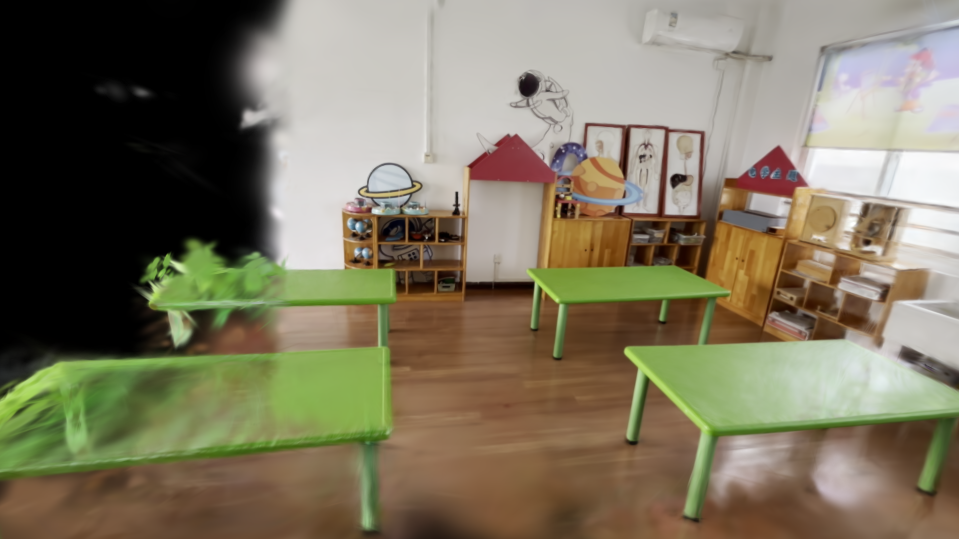} &
        \includegraphics[width=0.19\textwidth]{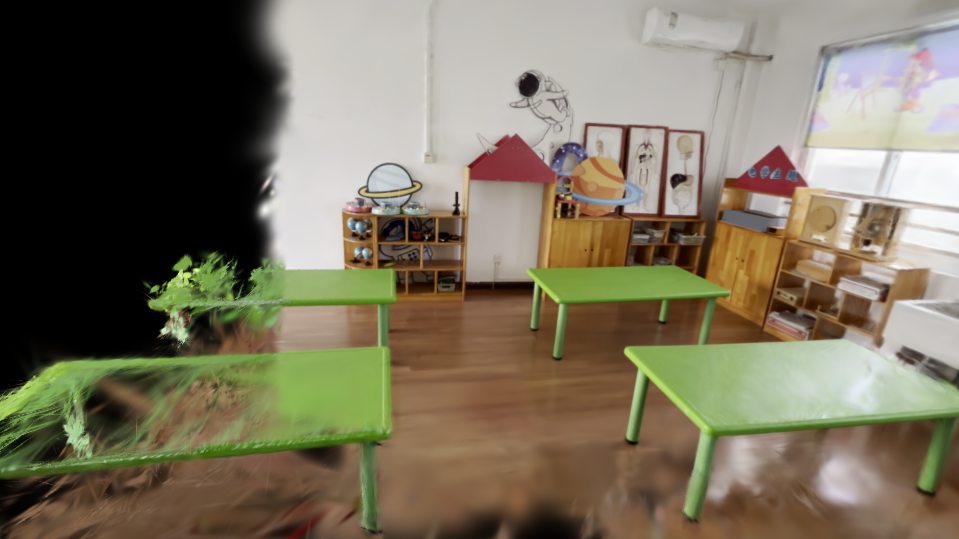} &
        \includegraphics[width=0.19\textwidth]{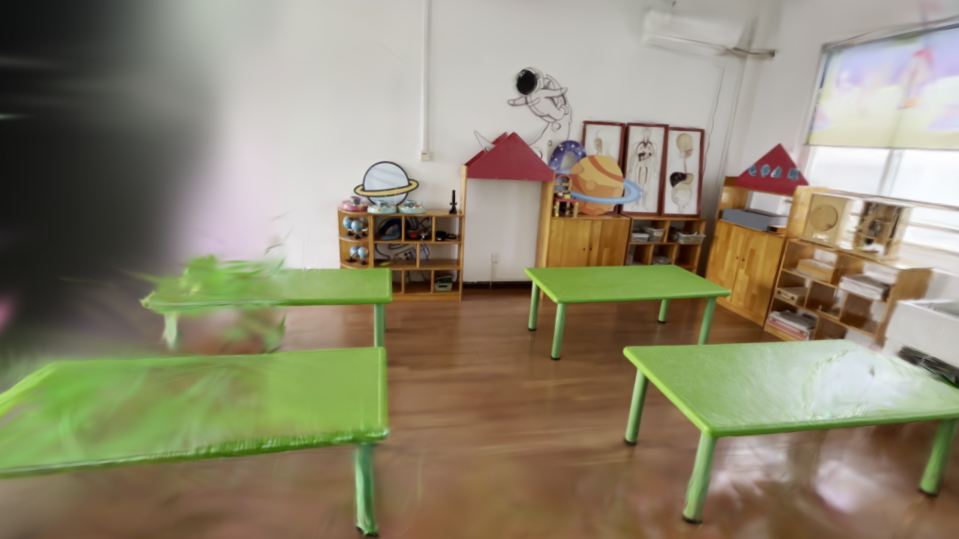} &
        \includegraphics[width=0.19\textwidth]{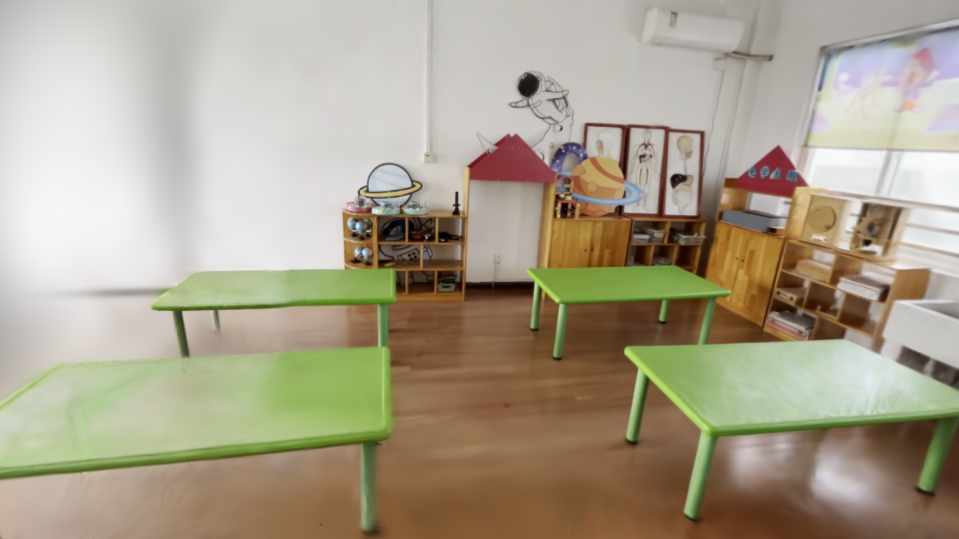} &
        \includegraphics[width=0.19\textwidth]{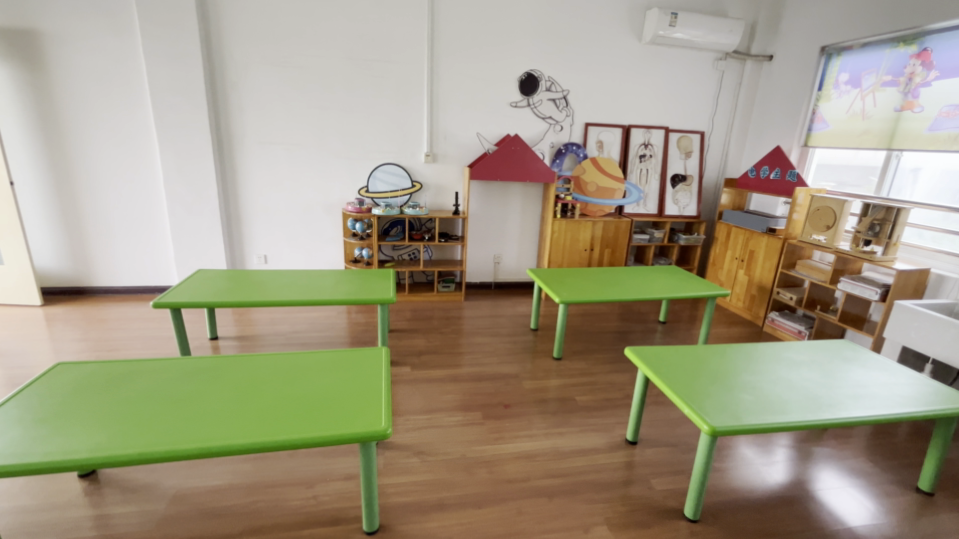} \\

        \includegraphics[width=0.19\textwidth]{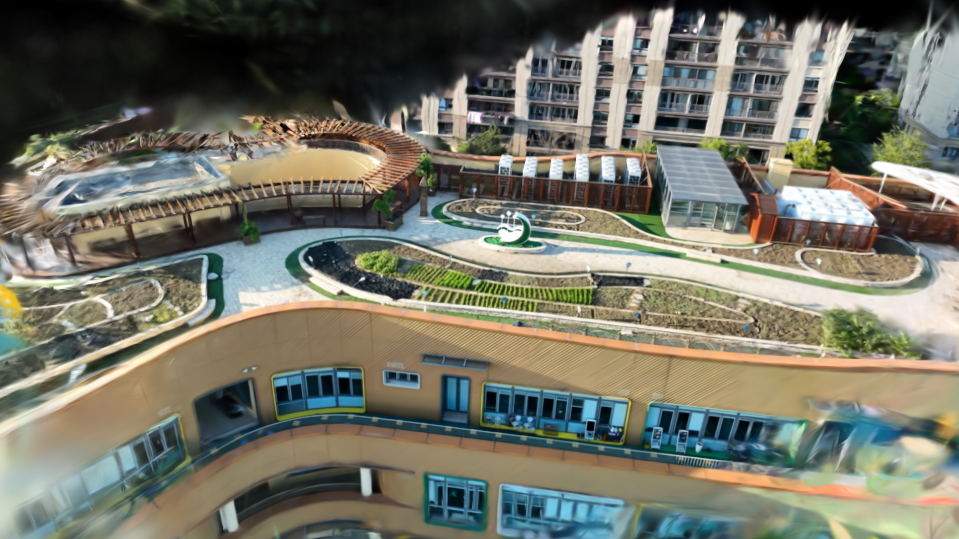} &
        \includegraphics[width=0.19\textwidth]{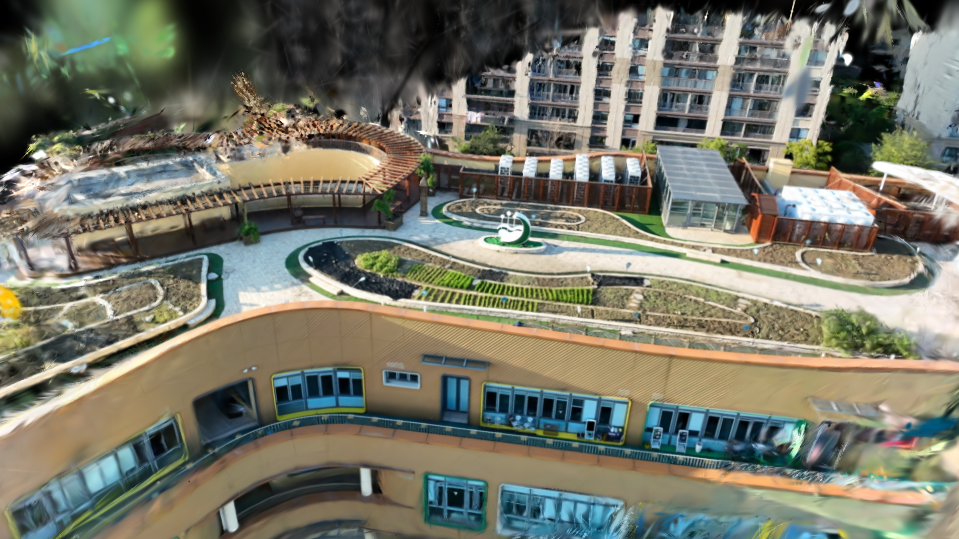} &
        \includegraphics[width=0.19\textwidth]{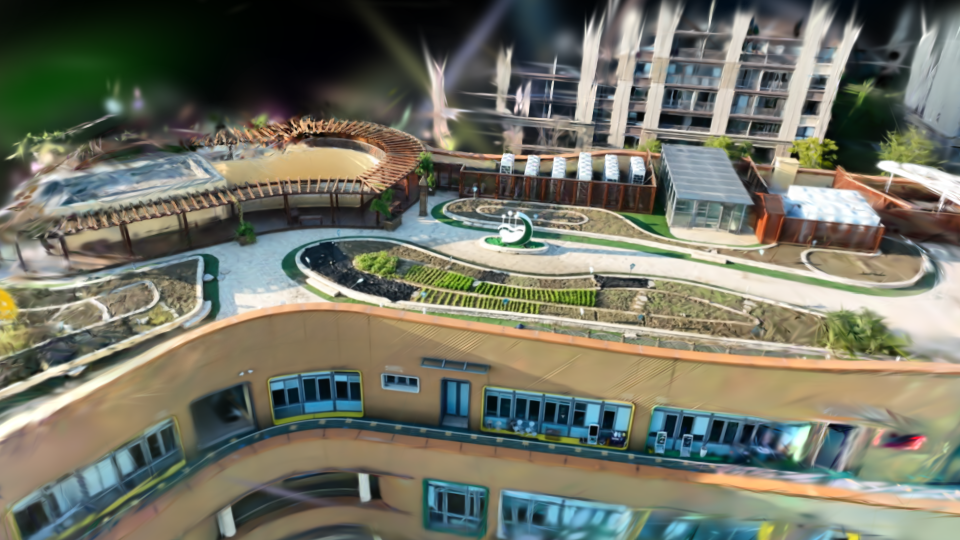} &
        \includegraphics[width=0.19\textwidth]{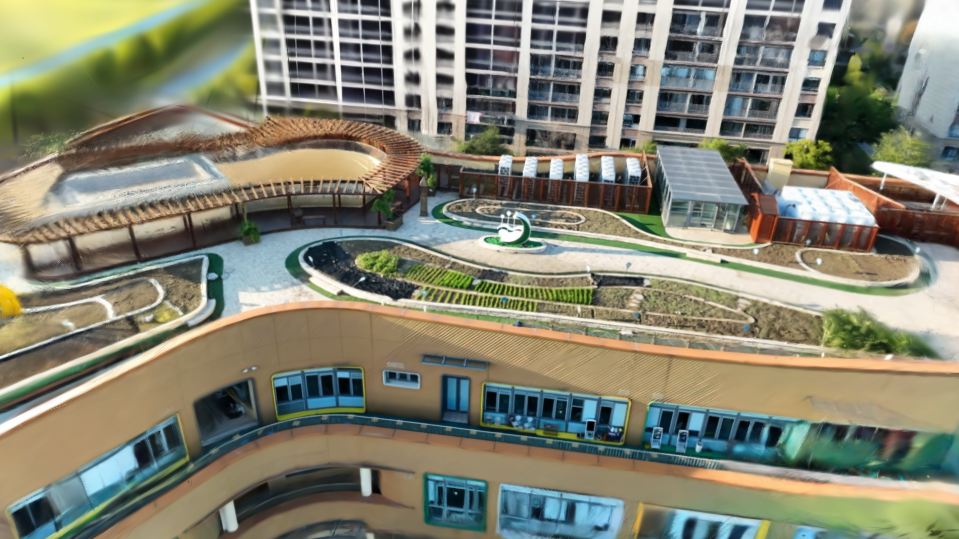} &
        \includegraphics[width=0.19\textwidth]{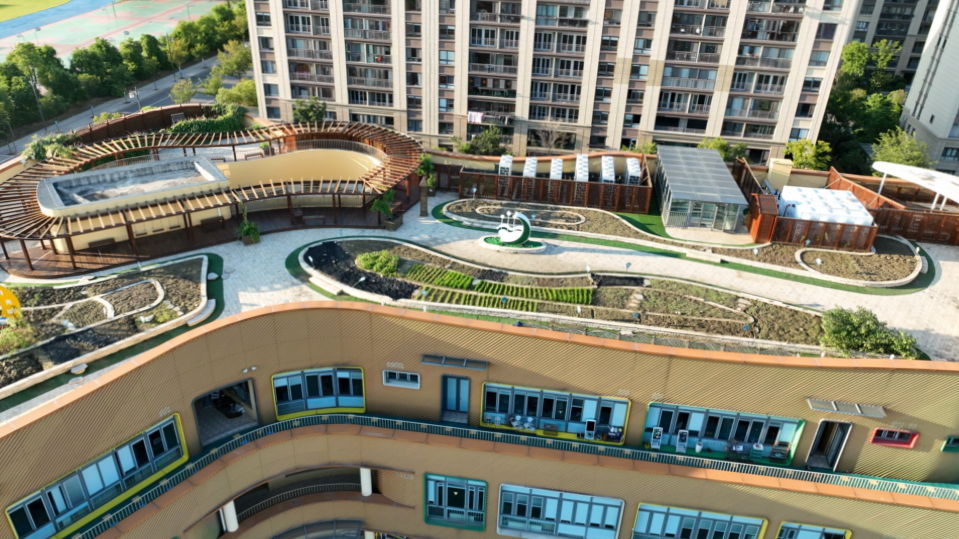} \\

        \includegraphics[width=0.19\textwidth]{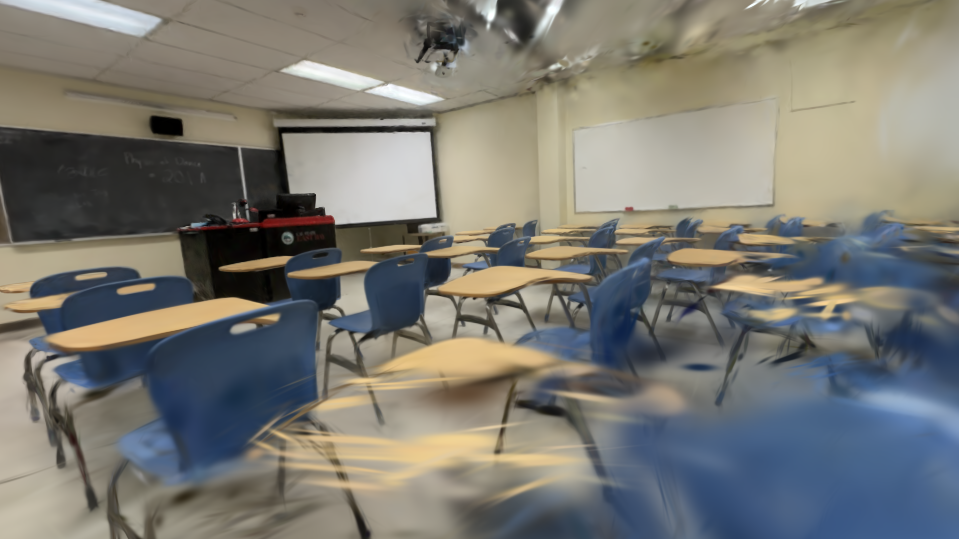} &
        \includegraphics[width=0.19\textwidth]{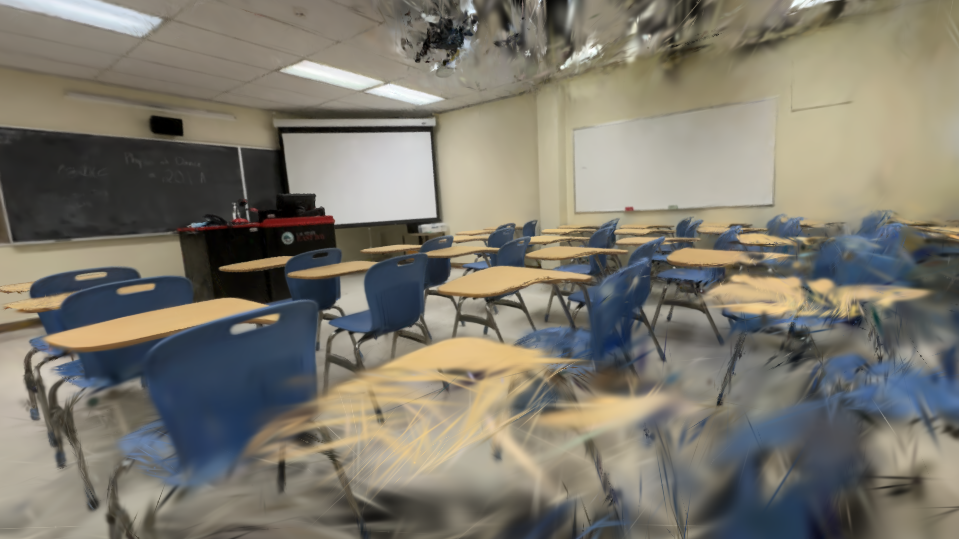} &
        \includegraphics[width=0.19\textwidth]{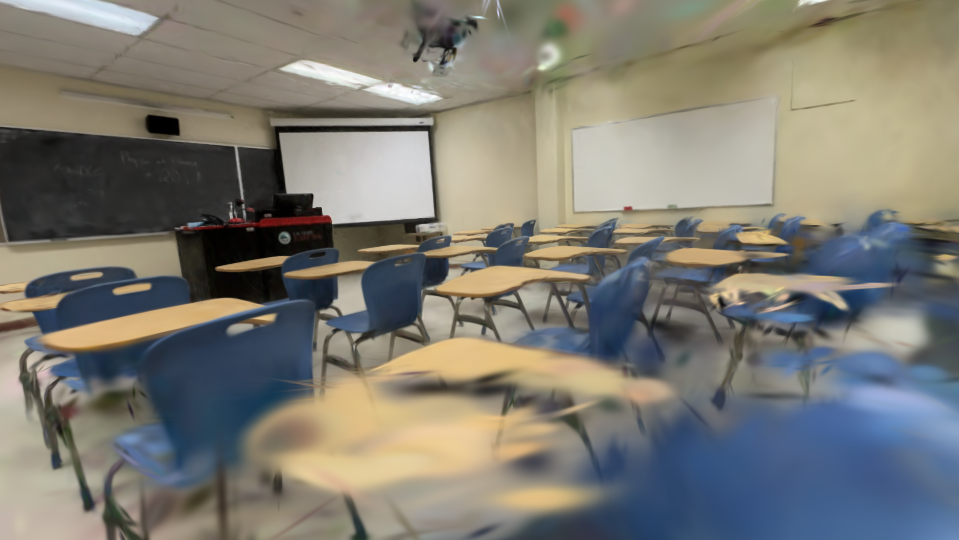} &
        \includegraphics[width=0.19\textwidth]{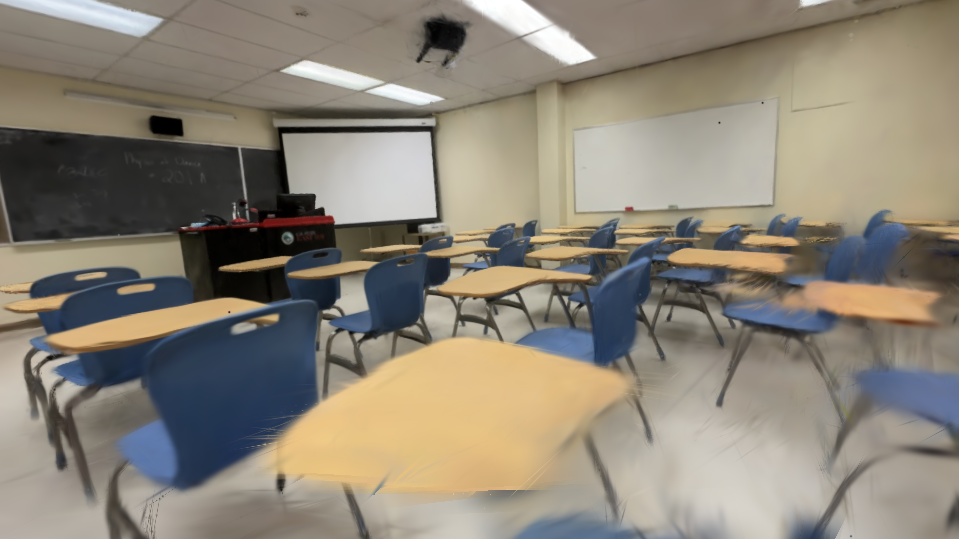} &
        \includegraphics[width=0.19\textwidth]{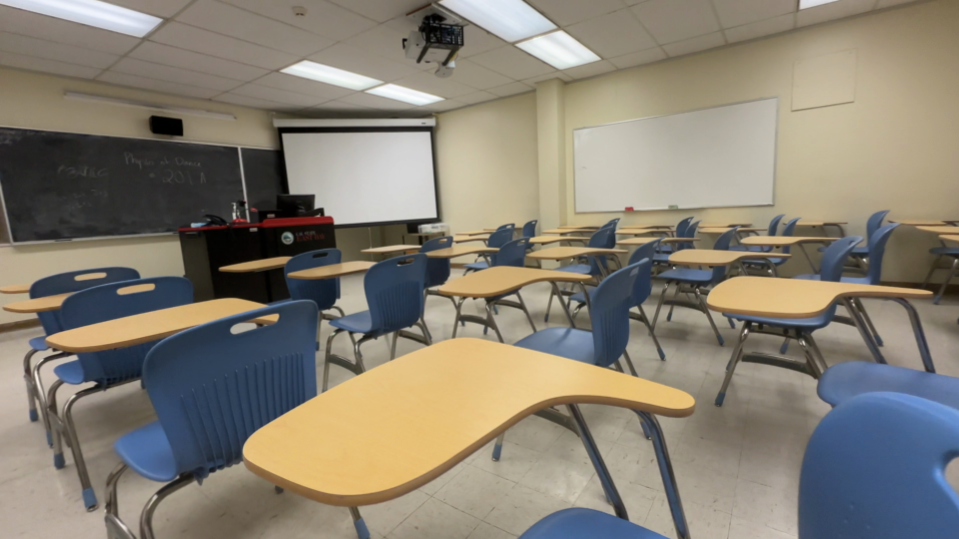} \\
        3DGS~\cite{Kerbl2023TOG} & 2DGS~\cite{Huang2DGS2024} & FSGS~\cite{Zehao2024ECCV} & Ours & GT
    \end{tabular}
    \vspace{-0.1in}
    \caption{\textbf{Qualitative comparison of novel view synthesis using masked input on DL3DV scenes \cite{DL3DV}.} Gaussian Splatting methods can easily overfit to training views, producing holes and needle-like artifacts when viewed from distant viewpoints. Our method effectively repairs these artifacts while augmenting invisible regions.}
    \label{fig:dl3dv_comparison}
    \vspace{-0.1in}
    
\end{figure*}

\subsection{View Interpolation}

Next, we evaluate our method in a view interpolation scenario where the target scene is fully covered by the input views, and testing views lie between these inputs— a common setup in prior regularization evaluations. We compare our method with previous regularization techniques on the Mip-NeRF 360 test set using 3, 6, and 9 input views, as shown in~\tabref{tab:sparse_view}.

Our results demonstrate that our method achieves more realistic novel view synthesis than the 3DGS and 2DGS baselines, as well as recent FSGS, by a significant margin.
It is worth noting that Gaussian Splatting is known to be more challenging to train than NeRF, especially in sparse view settings. We significantly narrow this gap, and, for the first time, show that Gaussian Splatting achieves performance comparable to state-of-the-art NeRF on the challenging Mip-NeRF360 dataset in sparse view settings.(see~\tabref{tab:sparse_view} and~\figref{fig:sparse_view}).


\begin{table}[]
    \renewcommand{\tabcolsep}{2pt}
    \centering
    \resizebox{1.0\linewidth}{!}{
    \begin{tabular}{l|ccc|ccc|ccc|ccc}
    & \multicolumn{3}{c|}{DL3DV $\nicefrac{1}{2}$ fps.} & \multicolumn{3}{c|}{TnT  $\nicefrac{1}{2}$ fps.} & \multicolumn{3}{c|}{DL3DV $\nicefrac{1}{4}$ fps.} & \multicolumn{3}{c}{TnT  $\nicefrac{1}{4}$ fps.}\\
    & PSNR$\uparrow$ & SSIM$\uparrow$ & LPIPS$\downarrow$ & PSNR$\uparrow$ & SSIM$\uparrow$ & LPIPS$\downarrow$ & PSNR$\uparrow$ & SSIM$\uparrow$ & LPIPS$\downarrow$ & PSNR$\uparrow$ & SSIM$\uparrow$ & LPIPS$\downarrow$ \\
    \hline
    
    3DGS~\cite{Kerbl2023TOG} &  \tbest17.22 &  \sbest0.740 & \sbest 0.314 & \tbest15.95 & \sbest0.653 & \sbest 0.414 & \tbest16.90 & \sbest0.728 & \sbest0.321 & \tbest14.75 & \sbest 0.609 & \best0.440 \\  
    2DGS~\cite{Huang2DGS2024} & 16.56 & 0.717 & \tbest0.323 & 15.46 &  \tbest0.640 & \best0.409 & 16.02 & 0.693 & \tbest0.336 & 14.38 & 0.589 & \best0.440 \\
    FSGS~\cite{Zhang2024CVPR} & \sbest18.25 & \tbest0.722 & 0.362 &  \sbest16.72 & 0.625 & 0.465& \sbest17.83 & \tbest0.710 & 0.370 & \sbest16.04 &\tbest 0.607 & \tbest0.473 \\
    Ours & \best 20.47 & \best 0.788 & \best0.284 & \best17.45 & \best0.662 & \tbest0.427 & \best20.01 & \best0.780 & \best0.292 & \best16.29 &\best 0.630 & \sbest0.447 
    \end{tabular}
    }
    \caption{\textbf{Quantitative comparison on DL3DV~\cite{DL3DV} and TnT~\cite{Knapitsch2017} datasets.} Each method is trained for 7,000 steps. Our method outperforms baselines by a significant margin.}
    \label{tab:crop_exp}
\end{table}


\subsection{View Extrapolation}
Unlike sparse view reconstruction, our paper focuses on a more practical scenario: generating complete, artifact-free 3D scenes from video captures. In reality, video offers dense sampling along the trajectory but highly sparse viewpoint coverage. While video is easy to capture and calibrate, it's challenging for 3D reconstruction, even reconstruction with state-of-the-art reconstruction methods can easily overfit to training views and produce serious artifacts for the viewpoints that are far away from the training views, see~\figref{fig:masked_recon}.
Therefore, we care about the novel view synthesis quality from the far field.

To this end, we propose a new evaluation protocol and reuse the masked reconstruction to mime the far-field rendering while preserving the reference, as introduced in~\secref{sec:generation}. Specifically, we downsample the video sequence at different time ratios for the train/test split (e.g., $\nicefrac{1}{2}$ and $\nicefrac{1}{4}$ in~\tabref{tab:crop_exp}) and select a patch covering about $25\%$ of the pixels in the training viewpoint for reconstruction, while evaluating on the full field of view of test viewpoints. Note that fixing the path location can result in many invisible and black regions, leading to bias in our generative methods, where the baselines may have significant errors in these areas. To ensure a reasonable comparison, we assign a trajectory to the masking path to provide more complete content coverage, see supplementary video.


\tabref{tab:crop_exp} and ~\figref{fig:dl3dv_comparison} show quantitative and qualitative results comparing our method with baselines on the DL3DV-benchmark and TnT scenes. GenFusion achieves significantly better rendering quality than the baselines, thanks to the strong prior from the diffusion model. It effectively removes needle-like artifacts and augments realistic 3D content for the invisible regions.

\boldstartspace{Scene Completion.} 
Furthermore, GenFusion not only provides content augmentation at the 3D scene boundary but also achieves scene-level completion, as shown in~\figref{fig:teaser}. Please refer to our webpage for more results.

\section{Discussion}
We have presented GenFusion, a novel model for artifact-free 3D asset generation. First, we adapt an existing well-trained 2D video diffusion model to drive a powerful 3D guidance with minimal modifications. Second, cyclical fusion enables scalable and robust 3D lifting, efficiently closing the loop between 3D reconstruction and generation through video synthesis.
Several limitations are evident: our method requiring additional denoising steps and slightly increasing training time (about 40 minutes per scene). Additionally, filling large invisible regions can cause blurriness due to inconsistency between video fragments. Modeling and addressing this inconsistency in the fusion module will be a key step toward achieving the next level of quality.

\boldparagraph{Acknowledgements}
This work was supported by the Westlake Education Foundation, supported by the Natural Science Foundation of Zhejiang province, China (No. QKWL25F0301), by the ERC Starting Grant LEGO-3D (850533), by the DFG EXC number 2064/1 - project number 390727645.

{
    \small
    \bibliographystyle{ieeenat_fullname}
    \bibliography{bibliography,bibliography_long,bibliography_custom}
}  

\clearpage
\setcounter{figure}{0}
\setcounter{table}{0}

\clearpage
\setcounter{page}{1}
\maketitlesupplementary
\appendix
\section{Overview}
In the supplementary materials, we provide comprehensive experimental details and extensive ablation studies to evaluate the contributions of our framework designs. Additionally, we present qualitative comparisons between our approach and baseline methods.

\section{Experimental Details}
\subsection{Video Diffusion Model Details}
Our diffusion model is built upon a pre-trained image-to-video latent diffusion model~\cite{dynamicrafter} which operates on RGB latent space.
However, we found that relying only on RGB inputs fails to produce consistent video frames, particularly in regions with severe artifacts. Therefore, we leverage depth maps to inject geometry information into the diffusion model.
To process RGB-D inputs, we utilize a pre-trained VAE from LDM3D~\cite{Gabriela2023}, which is designed to encode RGB-D image into the latent space. Therefore, given an RGB-D video of size $4{\times}T{\times}512{\times}320$ ($T$: video length), we flatten it along the first two dimensions, encode it into latent features of shape $4T{\times}64{\times}40$, and reshape to $4{\times}T{\times}64{\times}40$ for diffusion. For CLIP feature embedding, we randomly sample a reference frame from the input sequence. During reconstruction, the nearest input frame to the target trajectory serves as the reference for the CLIP guidance. In the training process, each training example comprises an artifact-prone RGB-D video, a reference image, and one target RGB-D video. To obtain the temporally consistent depth map for training, we leverage the SOTA monocular depth estimatior~\cite{hu2024-DepthCrafter} to augment the training data. During inference, we employ DDIM sampling with classifier-free guidance to modulate condition adherence strength. To do so, we implement random dropout of conditioning images with $10\%$ probability per sample during training.


In the video diffusion experiment section, we explore different designations of diffusion model to identify the optimal balance between model performance and computational efficiency. Therefore, four diffusion models are trained and analyzed in three aspects, input type, resolution, and video length. To this end, the base model that generates 16 frames of videos with a resolution of $512{\times}320$ is trained for 30k iterations using a learning rate of $1e-5$ and a batch size of 2 on each GPU. To assess the impact of depth information, we conduct a comparative analysis by training two base models: one utilizing RGB-D inputs and another with RGB inputs only. Both models are trained under identical hyperparameter settings to ensure a fair comparison. To enhance the quality of generated videos, we fine-tune the base RGBD model for higher resolution inputs (16 frames at $960{\times}512$) with an additional 34k iterations, maintaining the same learning rate and batch size configurations. To extend video generation capabilities, we fine-tune the temporal layers of our base model to produce 48-frame sequences for 30k iterations while maintaining the base model's batch size and learning rate.

\subsection{Masked 3D Reconstruction}
In the main paper, we introduce a masked 3D reconstruction scheme to mime the far-field rendering artifacts. The marked 3D reconstruction is used in both video diffusion data generation and novel view synthesis evaluation. In practice, we use a patch mask of size $H/2{\times}W/2$ to enable narrow field-of-view inputs in both settings. But differently, we randomly select one of the four corner locations for training dataset generation, since fixing the mask location introduces diverse artifacts and under-observed regions, enriching the dataset's complexity. 
However, the extremely limited observation setup often produces large black regions near the boundaries. Using such data directly for evaluation can lead to unrealistically low quantitative metrics in these regions due to content ambiguity.
To enable a fair comparison, we generate a trajectory to move the mask over time, rather than fixing its location as in video data generation. This ensures that most scene content is included in the input. Notably, all baselines and our method use the same sampling trajectories for each scene.
To further reduce sparsity along the camera trajectory, we downsample the viewpoints by factors of 2 and 4, using these masked frames as our training input while using the remaining full frames for evaluation.

\subsection{Cyclic Fusion}
We close the loop between reconstruction and generation through cyclic fusion that updates the 3D scene representation (\ie 2D Gaussian primitives) using input captures and generated videos.

\boldparagraph{Warm-up}
During the warm-up phase of the fusion process, the 3D representation is updated exclusively from input captures for the first 1000 iterations. Afterward, we apply our reconstruction-driven video diffusion every 1000 iterations to remove the artifacts and generate new content for the video renderings, which are then added to the training view set.

\boldparagraph{Sparsity-aware Densification}
In the original Gaussian Splatting~\cite{Kerbl2023TOG}, scene primitives are cloned and split based on the average magnitude of view-space position gradients, and the gradient for each primitive is reset every K steps (\ie, 100 steps in 3DGS and 2DGS). We find this strategy performs well in scenarios where the scene is densely captured. In such cases, primitives are typically observed for more than half of the reset steps ($>\frac{K}{2}$), making the averaged gradient over K steps a reliable indicator for deciding whether to add the primitive to the densification list.
However, this strategy becomes unreliable for masked 3D reconstruction, as the visibility counts of each Gaussian primitive are significantly lower, resulting in unstable gradient accumulation. To address this, we propose a sparsity-aware densification strategy that maintains the densification list by incorporating minimal visibility counts. Specifically, we disable gradient resets and add a primitive to the densification list only if its gradient exceeds the threshold and its visibility count surpasses the minimal visibility requirement. Accordingly, we perform the densification process every 100 iterations to progressively refine the point cloud representation. We found this strategy is more robust for handling diverse input scenarios.

\begin{figure*}
    \centering
    \includegraphics[width=0.95\linewidth]{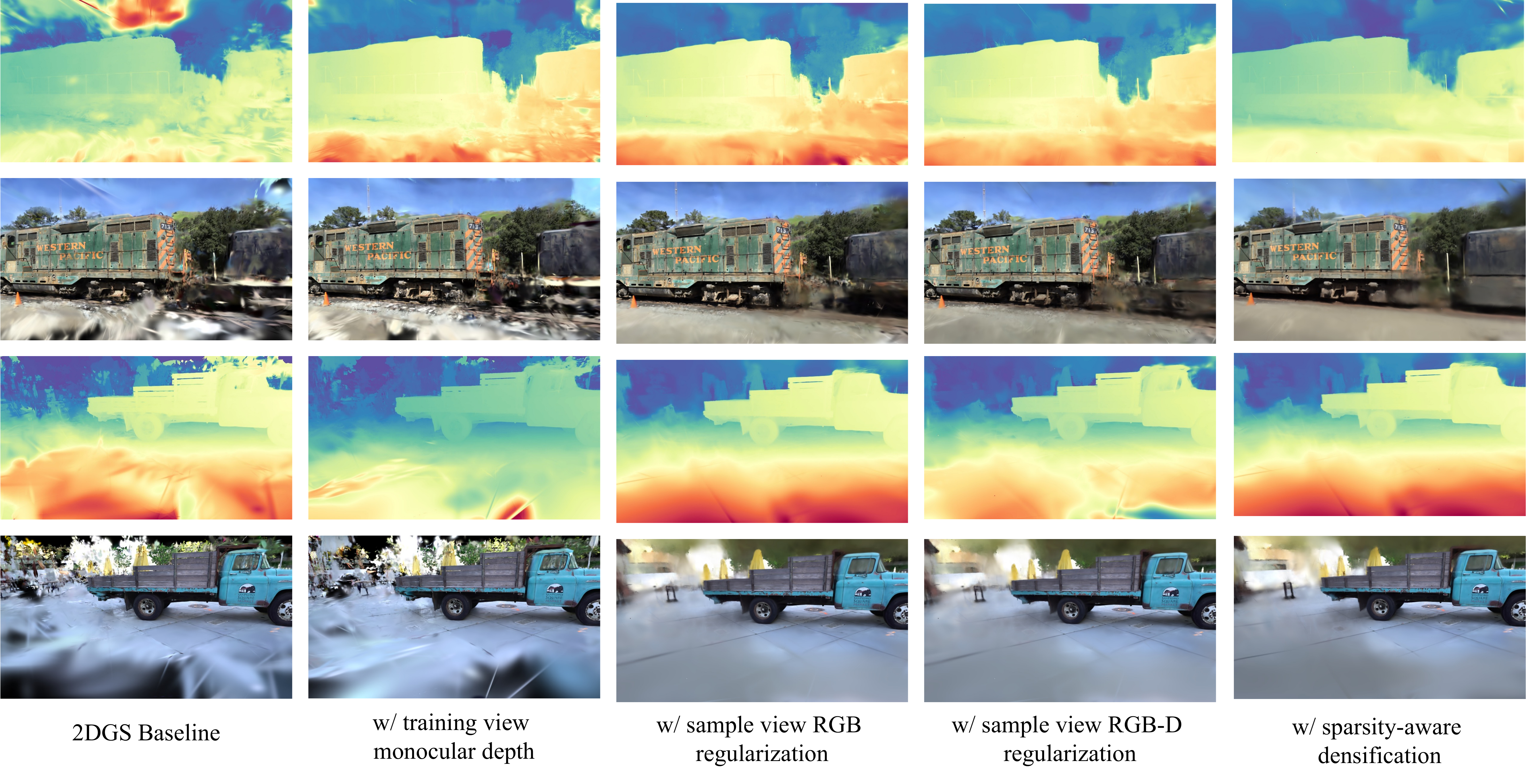}
    \caption{From top to bottom: 2DGS baseline, with train view monocular depth added, with sample view RGB added, with sample view depth added, and finally with sparsity-aware densification.}
    \label{fig:ablations}
    \vspace{-0.1in}
\end{figure*}

\section{Ablation Studies}
In \tabref{tab:ablation}, we perform comprehensive ablation studies to validate the contributions of our model components using scenes from the Tanks and Temples dataset\cite{Knapitsch2017}. We begin with a vanilla 2D Gaussian Splatting (2DGS) model, following its original implementation, as the baseline. Building on this, we evaluate the effect of incorporating monocular depth supervision during training and view sampling using the ScaleAndShiftInvariant loss~\cite{Ranftl2020PAMI}. As shown in (2) of ~\tabref{tab:ablation}, this addition does not yield quantitative improvements. However, it encourages smoothness in the rendered depth, effectively reducing floating artifacts typically observed during initial reconstruction stages (visualized in ~\figref{fig:ablations}).
Significant performance gains are observed in (3) and (5) of ~\tabref{tab:ablation}, attributed to our RGB regularization and sparsity-aware densification strategies, further confirming the effectiveness of our method.


\begin{table}[t]
    \centering
    \resizebox{\linewidth}{!}{
        \begin{tabular}{l|cccc}
            \hline
            \makebox[0.01\textwidth][c]{No.} & \makebox[0.03\textwidth][c]{Method} & PSNR$\uparrow$ & SSIM$\uparrow$ & LPIPS$\downarrow$ \\
            \hline
            1 & 2DGS baseline & 13.87 & 0.572 & 0.447 \\
            2 & +train view monocular depth & 13.89 & 0.575 & 0.442 \\
            3 & +sample view rgb & 15.33 & 0.602 & 0.442 \\
            4 & +sample view depth & 15.34 & 0.622 & 0.438 \\
            5 & +sparsity aware densification & 15.81 & 0.617 & 0.409 \\
            \hline
        \end{tabular}
    }
    \caption{Ablation studies using on Tanks and Temples dataset. $\uparrow$ indicates higher is better, while $\downarrow$ indicates lower is better.}
    \label{tab:ablation}
\end{table}

\section{More Evaluation}
\subsection{View Interpolation}
\tabref{tab:per-scene} provides a per-scene break down for quantity metrics in Mip-NeRF360. These results showcase that our models consistently improve the baselines.
%

\subsection{View Extrapolation and Scene Completion}
\begin{figure*}[htbp]
    \vspace{-0.2in}
    \centering
    \setlength{\tabcolsep}{0.2em}
      \renewcommand{\arraystretch}{0.4}
    \begin{tabular}{cccccc}
        \includegraphics[width=0.19\textwidth]{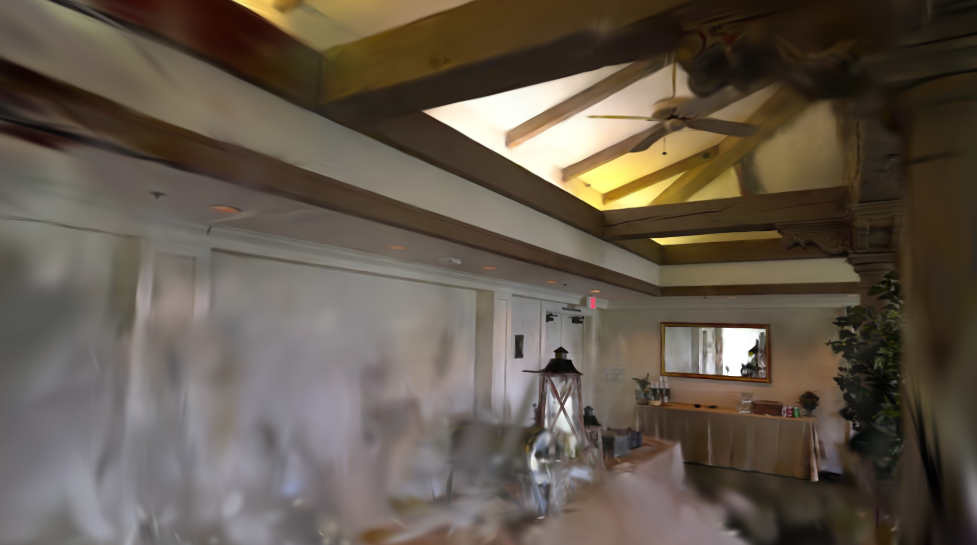} &
        \includegraphics[width=0.19\textwidth]{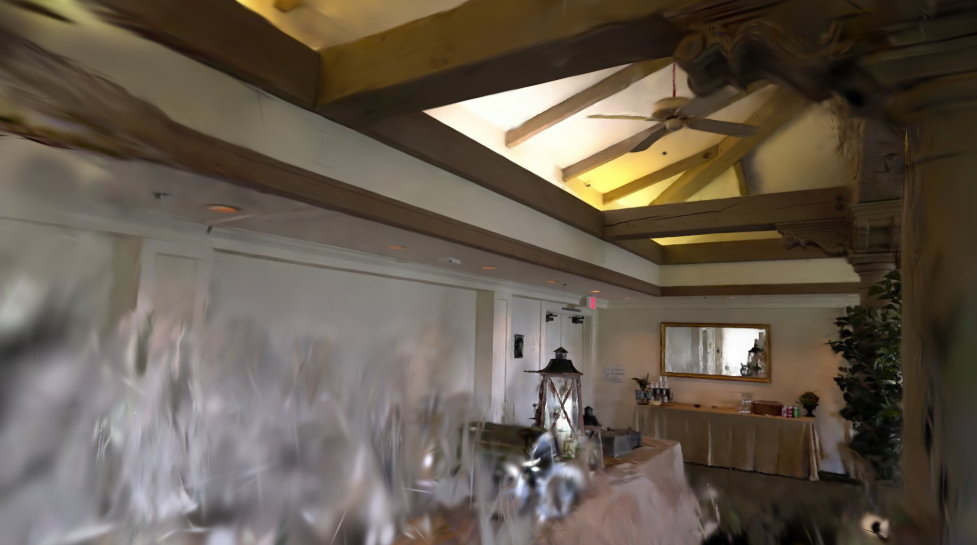} &
        \includegraphics[width=0.19\textwidth]{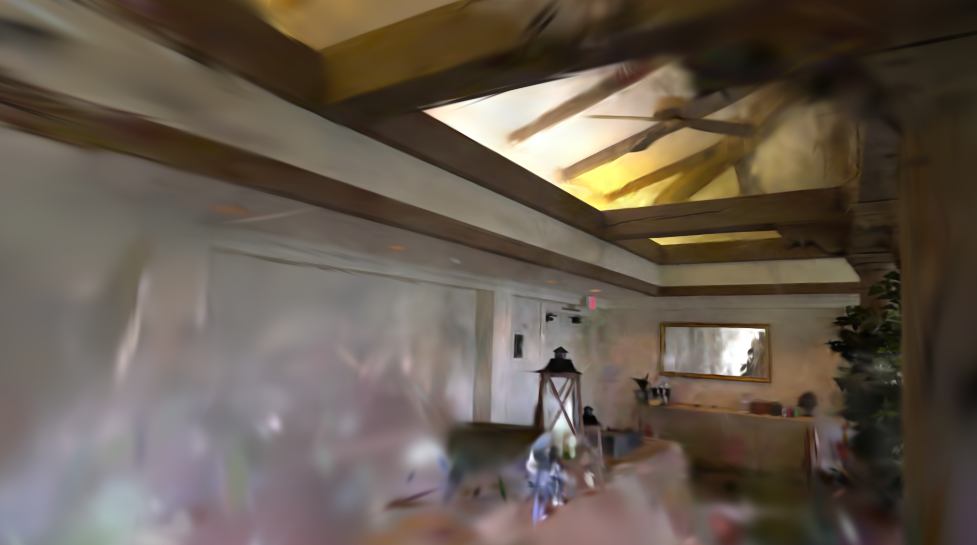} &
        \includegraphics[width=0.19\textwidth]{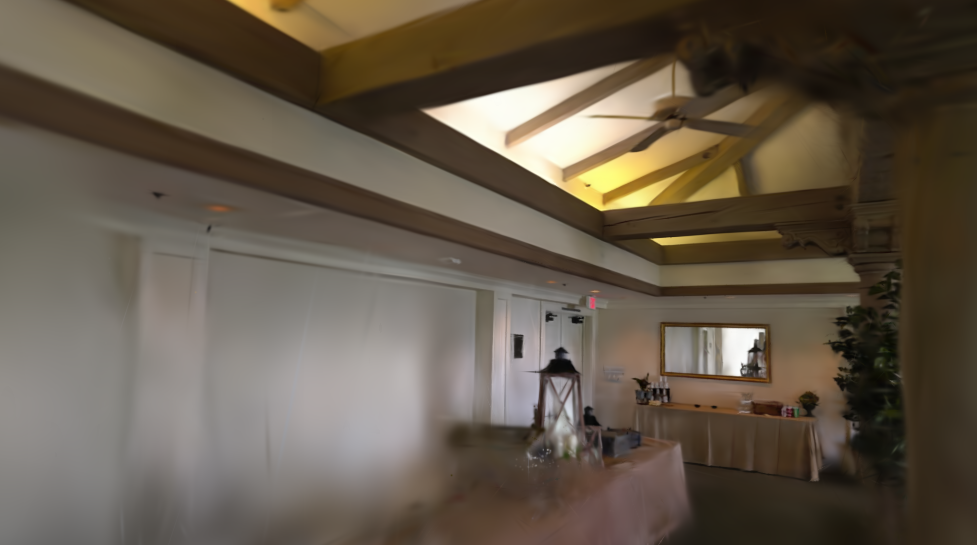} &
        \includegraphics[width=0.19\textwidth]{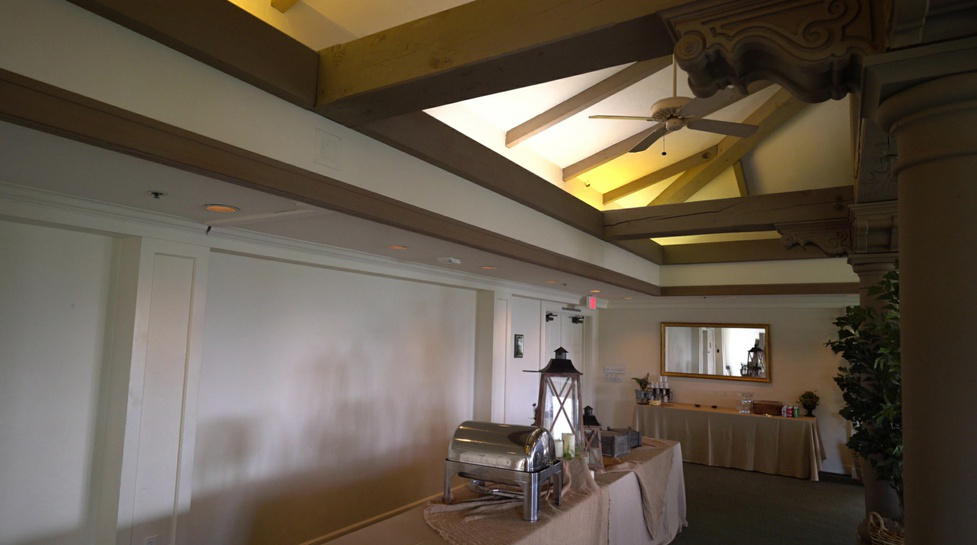} \\
        
        \includegraphics[width=0.19\textwidth]{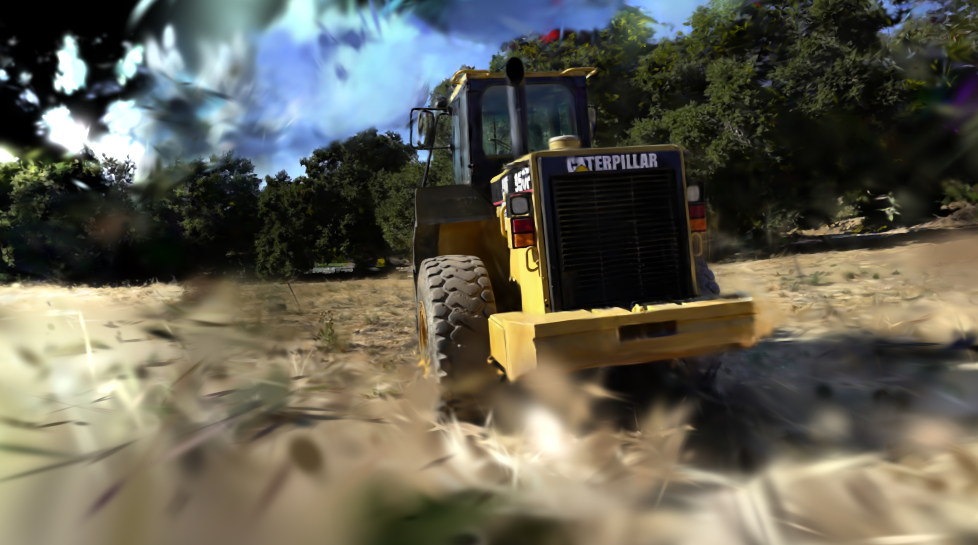} &
        \includegraphics[width=0.19\textwidth]{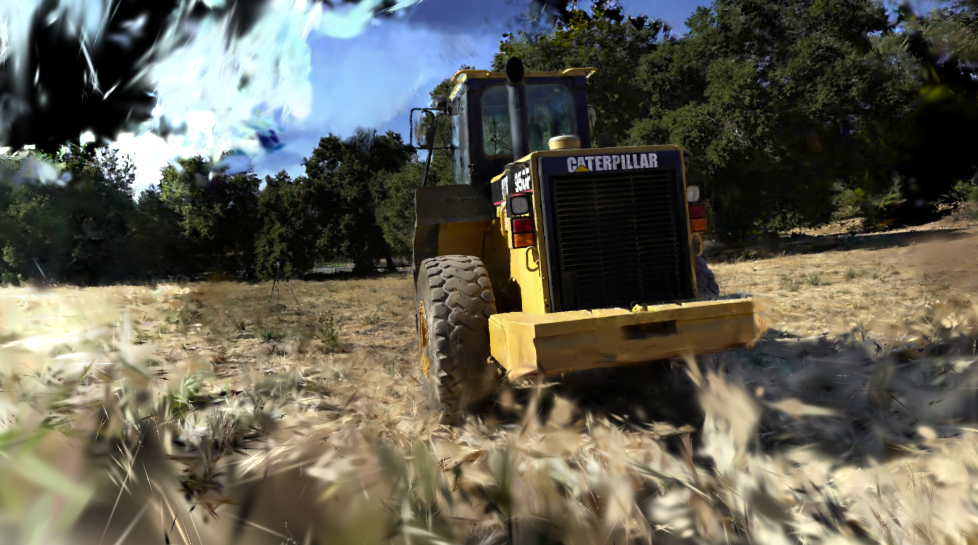} &
        \includegraphics[width=0.19\textwidth]{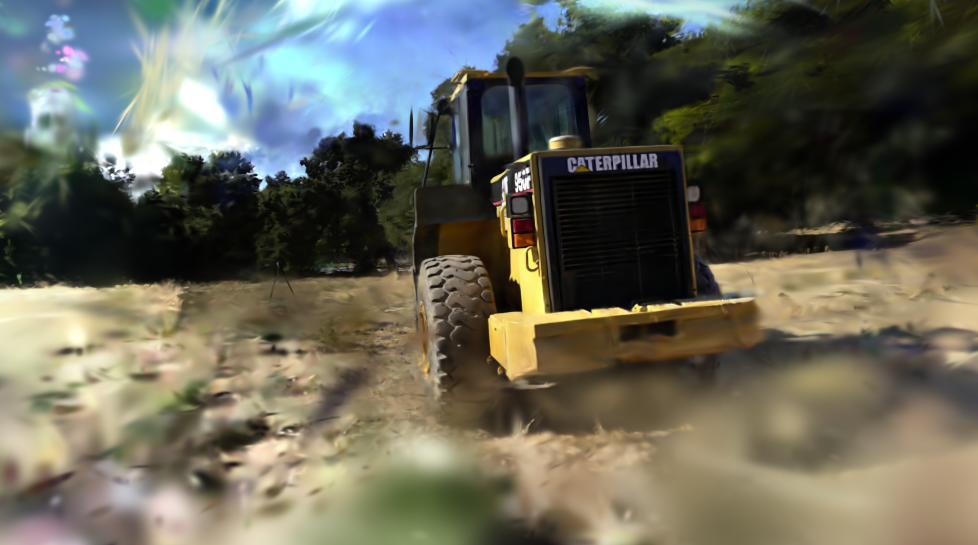} &
        \includegraphics[width=0.19\textwidth]{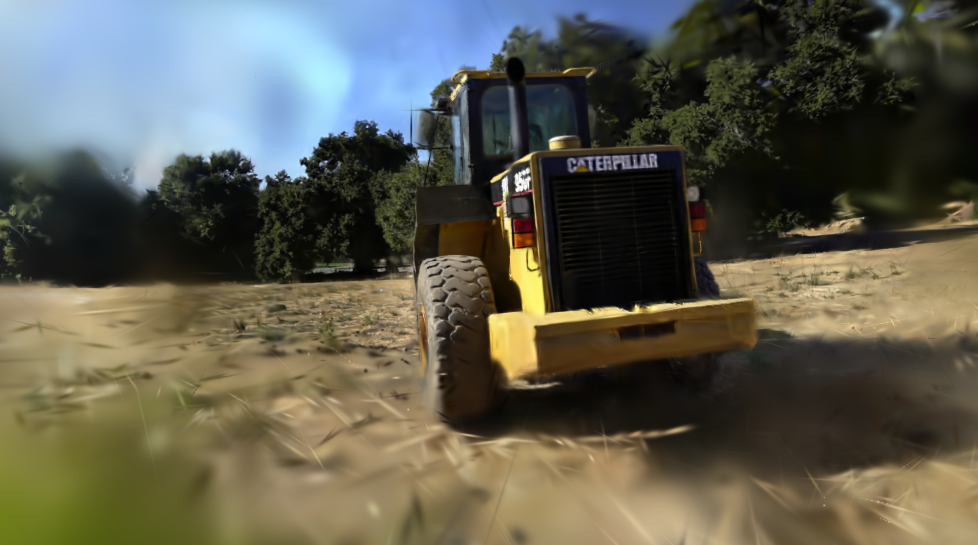} &
        \includegraphics[width=0.19\textwidth]{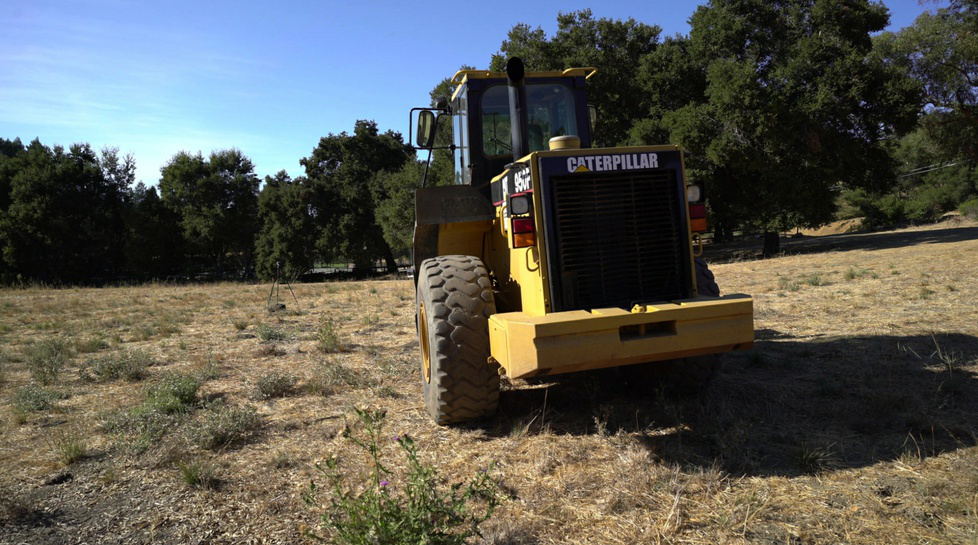} \\

        \includegraphics[width=0.19\textwidth]{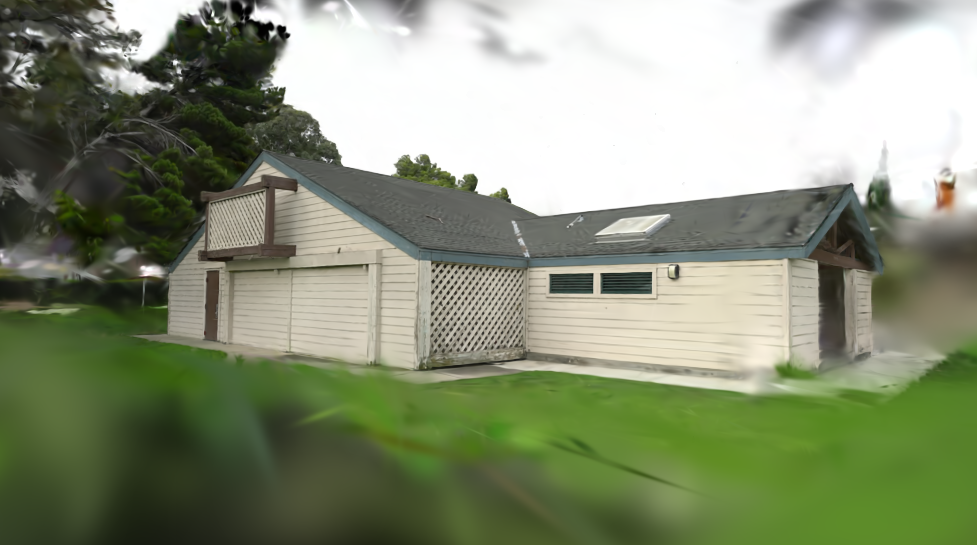} &
        \includegraphics[width=0.19\textwidth]{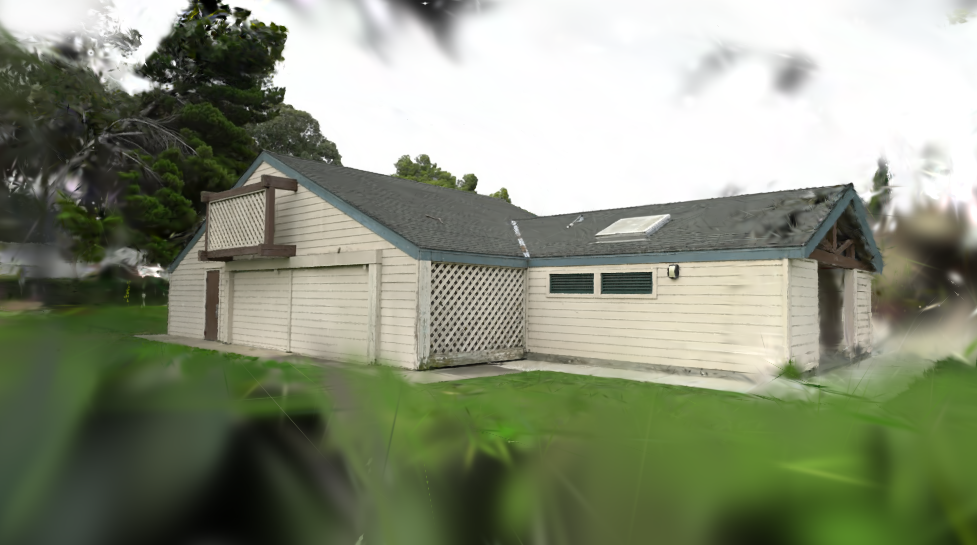} &
        \includegraphics[width=0.19\textwidth]{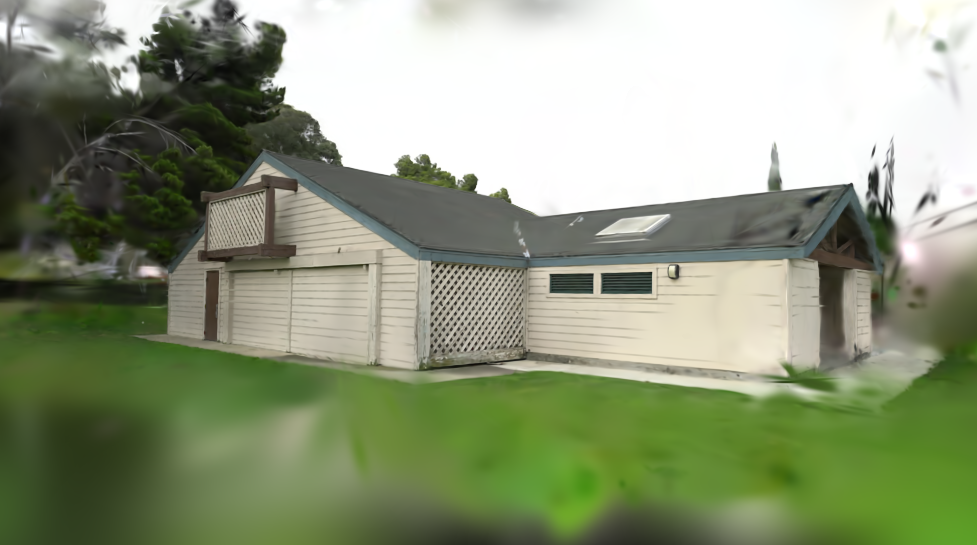} &
        \includegraphics[width=0.19\textwidth]{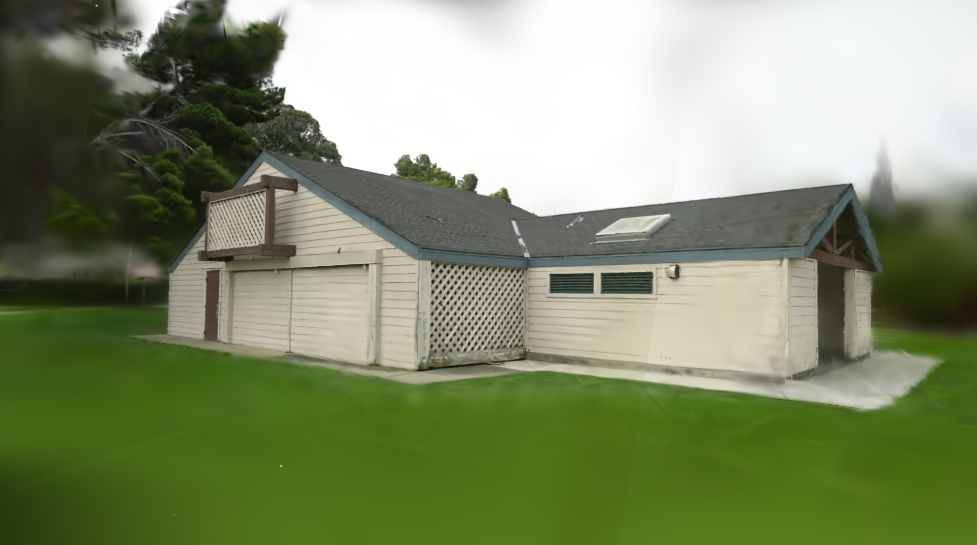} &
        \includegraphics[width=0.19\textwidth]{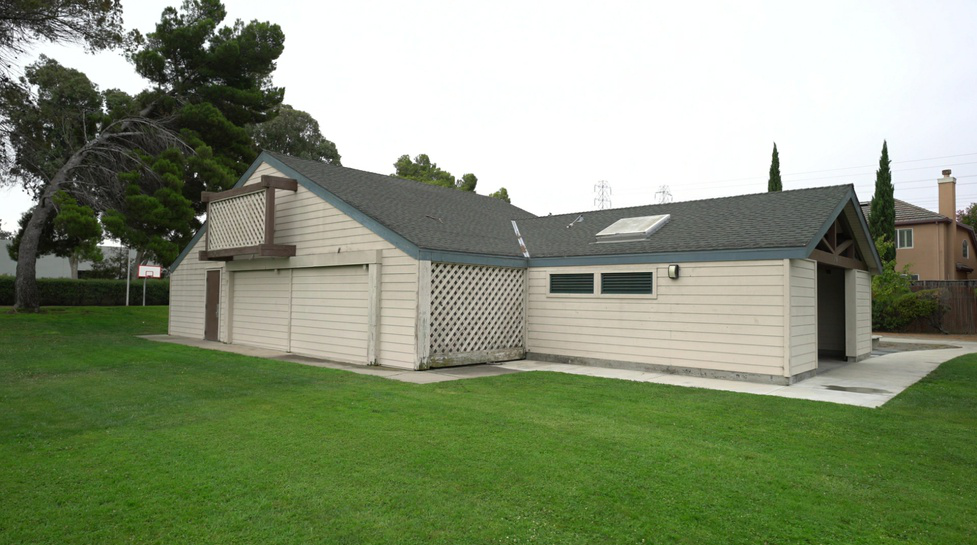} \\

        3DGS~\cite{Kerbl2023TOG} & 2DGS~\cite{Huang2DGS2024} & FSGS~\cite{Zehao2024ECCV} & Ours & GT
    \end{tabular}
    \vspace{-0.1in}
    \caption{Qualitative comparison of novel view synthesis using masked input on TnT scenes~\cite{Knapitsch2017}.}
    \label{fig:tnt_comparison}
    \vspace{-0.1in}
    
\end{figure*}

Here we present extensive experimental results on masked 3D reconstruction. \figref{fig:tnt_comparison} demonstrate that our performance also outperforms baselines in far-field viewpoint renderings. ~\tabref{tab:crop_exp_per_scene_tnt} and ~\tabref{tab:crop_exp_per_scene_dl3dv} provide per-scene quantitative results.

\section{Conclusion}
We have observed viewpoint saturation as a fundamental limitation in previous reconstruction and generation methods: high-quality reconstruction relies on dense captures, while generation methods are optimized for weak conditioning.
To relax this constraint, we propose GenFusion, an efficient generative guidance framework that enables accurate 3D reconstruction and content generation for input conditions across varying densities. 
We achieve this by closing the loop between reconstruction and generation, creating a feedback loop where generation becomes aware of the reconstruction status through novel trajectory rendering, and reconstruction is further regularized using RGB-D videos generated by our video diffusion model.
We evaluate the interpolation capability using a sparse view reconstruction setup and the extrapolation capability with a novel masked reconstruction mechanism. Both tasks demonstrate significant improvements over baseline methods. In addition, our approach achieves scene-level 3D completion, enabling 3D scene expansion.
We hope our findings in bridging reconstruction and generation can inspire other novel view syntheses and 3D scene generation tasks.

\begin{table*}[t]
    \renewcommand{\tabcolsep}{3.5pt}
    \centering
    \begin{tabular}{l|ccc|ccc|ccc|ccc}
    & \multicolumn{3}{c|}{2DGS} & \multicolumn{3}{c|}{3DGS } & \multicolumn{3}{c}{FSGS}& \multicolumn{3}{c}{Ours}\\
    & PSNR$\uparrow$ & SSIM$\uparrow$ & LPIPS$\downarrow$ & PSNR$\uparrow$ & SSIM$\uparrow$ & LPIPS$\downarrow$ & PSNR$\uparrow$ & SSIM$\uparrow$ & LPIPS$\downarrow$ & PSNR$\uparrow$ & SSIM$\uparrow$ & LPIPS$\downarrow$ \\
    \hline
        14eb48a50e  & 16.44 & 0.690 & 0.384 & 17.40 & 0.730 & 0.360 & 17.64 & 0.690 & 0.434 & 19.92 & 0.767 & 0.351\\
        0a1b7c20a9  & 15.74 & 0.733 & 0.278 & 16.38 & 0.760 & 0.263 & 18.17 & 0.752 & 0.310 & 19.27 & 0.811 & 0.230\\
        06da796666  & 15.42 & 0.672 & 0.396 & 15.34 & 0.698 & 0.390 & 17.02 & 0.710 & 0.437 & 18.54 & 0.755 & 0.376\\
        389a460ca1  & 18.04 & 0.810 & 0.309 & 18.24 & 0.824 & 0.311 & 18.48 & 0.799 & 0.367 & 21.11 & 0.861 & 0.273\\
        2cbfe28643  & 16.09 & 0.782 & 0.257 & 16.79 & 0.799 & 0.254 & 19.50 & 0.790 & 0.321 & 22.03 & 0.850 & 0.227\\
        374ffd0c5f  & 19.85 & 0.780 & 0.256 & 21.16 & 0.803 & 0.250 & 20.98 & 0.763 & 0.327 & 22.35 & 0.842 & 0.224\\
        5c3af58102  & 15.66 & 0.692 & 0.273 & 15.95 & 0.709 & 0.260 & 16.22 & 0.661 & 0.325 & 20.10 & 0.794 & 0.214\\
        66fd66cbed  & 21.42 & 0.855 & 0.235 & 22.15 & 0.873 & 0.224 & 22.29 & 0.867 & 0.246 & 23.27 & 0.897 & 0.191\\
        3bb3bb4d3e  & 16.89 & 0.795 & 0.266 & 17.85 & 0.810 & 0.253 & 18.84 & 0.780 & 0.319 & 22.48 & 0.883 & 0.198\\
        91afb9910b  & 19.18 & 0.765 & 0.274 & 19.91 & 0.773 & 0.278 & 20.86 & 0.776 & 0.304 & 22.76 & 0.820 & 0.240\\
        7705a2edd0  & 16.74 & 0.698 & 0.398 & 16.78 & 0.712 & 0.396 & 18.89 & 0.715 & 0.440 & 21.71 & 0.792 & 0.350\\
        71b2dc8a2a  & 15.67 & 0.796 & 0.264 & 15.94 & 0.814 & 0.252 & 20.42 & 0.857 & 0.252 & 21.64 & 0.887 & 0.199\\
        a726c1112a  & 18.60 & 0.804 & 0.321 & 19.45 & 0.832 & 0.295 & 17.02 & 0.726 & 0.423 & 20.00 & 0.83 & 0.297\\
        cbd44beb04  & 16.46 & 0.700 & 0.311 & 17.40 & 0.728 & 0.299 & 17.35 & 0.706 & 0.349 & 19.38 & 0.789 & 0.285\\
        df4f9d9a0a  & 17.21 & 0.743 & 0.358 & 18.04 & 0.768 & 0.344 & 19.36 & 0.777 & 0.356 & 21.82 & 0.845 & 0.268\\
        6d22162561  & 15.79 & 0.663 & 0.398 & 16.62 & 0.681 & 0.401 & 18.07 & 0.671 & 0.441 & 20.58 & 0.737 & 0.372\\
        6d81c5ab0d  & 13.19 & 0.540 & 0.448 & 14.22 & 0.601 & 0.425 & 14.47 & 0.573 & 0.492 & 16.42 & 0.634 & 0.448\\
        ec305787b7  & 16.75 & 0.751 & 0.286 & 16.98 & 0.763 & 0.278 & 16.78 & 0.688 & 0.381 & 22.122 & 0.846 & 0.211\\
        85cd0e9211  & 18.17 & 0.758 & 0.285 & 18.45 & 0.767 & 0.289 & 18.90 & 0.688 & 0.372 & 22.73 & 0.814 & 0.269\\
        95e4b24092  & 13.97 & 0.581 & 0.353 & 13.66 & 0.596 & 0.351 & 14.99 & 0.598 & 0.373 & 15.99 & 0.609 & 0.345\\
        7da3db9905  & 16.51 & 0.737 & 0.309 & 18.69 & 0.778 & 0.285 & 19.98 & 0.765 & 0.314 & 22.09 & 0.831 & 0.231\\
        d3812aad53  & 15.09 & 0.607 & 0.454 & 16.16 & 0.654 & 0.438 & 16.80 & 0.662 & 0.451 & 17.43 & 0.684 & 0.428\\
        b0c4613d6c  & 15.10 & 0.612 & 0.332 & 15.54 & 0.623 & 0.336 & 17.71 & 0.637 & 0.368 & 19.00 & 0.668 & 0.324\\
        b4f53094fd  & 13.48 & 0.634 & 0.306 & 14.28 & 0.653 & 0.299 & 17.17 & 0.677 & 0.311 & 18.59 & 0.698 & 0.271\\
        average  & 16.56 & 0.717 & 0.323 & 17.22 & 0.740 & 0.314 & 18.25 & 0.722 & 0.363 & 20.47 & 0.788 & 0.284 \\
        \hline
   
    \end{tabular}
    \caption{Quantitative comparison on DL3DV datasets. Each method is trained on 7000 steps.}
    \label{tab:crop_exp_per_scene_dl3dv}
\end{table*}

\begin{table*}[ht]
    \renewcommand{\tabcolsep}{3.5pt}
    \centering
    \begin{tabular}{l|ccc|ccc|ccc|ccc}
    & \multicolumn{3}{c|}{2DGS} & \multicolumn{3}{c|}{3DGS } & \multicolumn{3}{c}{FSGS}& \multicolumn{3}{c}{Ours}\\
    & PSNR$\uparrow$ & SSIM$\uparrow$ & LPIPS$\downarrow$ & PSNR$\uparrow$ & SSIM$\uparrow$ & LPIPS$\downarrow$ & PSNR$\uparrow$ & SSIM$\uparrow$ & LPIPS$\downarrow$ & PSNR$\uparrow$ & SSIM$\uparrow$ & LPIPS$\downarrow$ \\
\hline
barn  & 16.80 & 0.675 & 0.371 & 17.64 & 0.685 & 0.377 & 18.47 & 0.677 & 0.405 & 17.84 & 0.672 & 0.402\\
ignatius  & 15.75 & 0.588 & 0.329 & 15.88 & 0.591 & 0.359 & 16.14 & 0.521 & 0.458  & 17.51 & 0.614 & 0.363\\
meetingroom  & 17.63 & 0.672 & 0.364 & 17.80 & 0.694 & 0.356 & 17.71 & 0.667 & 0.421 & 19.37 & 0.733 & 0.348\\
truck  & 14.66 & 0.646 & 0.357 & 15.39 & 0.663 & 0.361 & 16.69 & 0.654 & 0.407 & 16.80 & 0.673 & 0.383\\
courthouse  & 14.80 & 0.630 & 0.411 & 15.15 & 0.640 & 0.419 & 15.80 & 0.632 & 0.454 & 15.68 & 0.622 & 0.461\\
caterpillar  & 13.79 & 0.532 & 0.403 & 14.33 & 0.542 & 0.431 & 15.35 & 0.530 & 0.490& 16.58 & 0.580 & 0.432 \\
train  & 13.77 & 0.561 & 0.423 & 14.31 & 0.587 & 0.424 & 14.47 & 0.528 & 0.516& 15.34 & 0.580 & 0.458\\
average  & 15.31 & 0.615 & 0.380 & 15.79 & 0.629 & 0.390 & 16.38 & 0.601 & 0.450 & 17.01 & 0.639 &  0.406\\
\hline

    \end{tabular}
    \caption{Quantitative comparison on TnT datasets. Each method is trained on 7000 steps with 1/2 frames}
    \label{tab:crop_exp_per_scene_tnt}
\end{table*}

\begin{table*}[ht]
    \renewcommand{\tabcolsep}{4.5pt}
    \centering
    \begin{tabular}{l|ccc|ccc|ccc|ccc}
    & \multicolumn{3}{c|}{2DGS} & \multicolumn{3}{c|}{3DGS } & \multicolumn{3}{c|}{FSGS}& \multicolumn{3}{c}{Ours}\\
    & PSNR$\uparrow$ & SSIM$\uparrow$ & LPIPS$\downarrow$ & PSNR$\uparrow$ & SSIM$\uparrow$ & LPIPS$\downarrow$ & PSNR$\uparrow$ & SSIM$\uparrow$ & LPIPS$\downarrow$ & PSNR$\uparrow$ & SSIM$\uparrow$ & LPIPS$\downarrow$ \\
\hline
\multicolumn{13}{c}{3 Views} \\
\hline
bicycle  & 12.70 & 0.124 & 0.622 & 14.33 & 0.300 & 0.556 & 14.30 & 0.234 & 0.624 & 15.46 & 0.275 & 0.647\\
bonsai  & 11.60 & 0.300 & 0.568 & 10.92 & 0.301 & 0.736 & 13.75 & 0.376 & 0.524 & 14.12 & 0.418 & 0.534\\
counter  & 13.17 & 0.311 & 0.539 & 12.62 & 0.305 & 0.597 & 13.99 & 0.392 & 0.527 & 15.20 & 0.470 & 0.520\\
garden  & 13.06 & 0.184 & 0.575 & 12.08 & 0.145 & 0.649 & 14.33 & 0.274 & 0.586 & 16.65 & 0.305 & 0.580\\
room  & 13.79 & 0.410 & 0.490 & 13.04 & 0.342 & 0600 & 14.26 & 0.483 & 0.484 & 16.40 & 0.570 & 0.438\\
stump  & 14.63 & 0.171 & 0.593 & 14.10 & 0.196 & 0.626 & 15.93 & 0.276 & 0.607 & 17.13 & 0.317 & 0.640 \\
kitchen  & 14.07 & 0.307 & 0.542 & 13.35 & 0.257 & 0.621 & 14.76 & 0.361 & 0.538 & 16.02 & 0.427 & 0.542\\
flowers  & 10.57 & 0.104 & 0.657 & 10.08 & 0.129 & 0.794 & 12.17 & 0.177 & 0.664 & 12.89 & 0.210 & 0.715\\
treehill & 11.95 & 0.186 & 0.627 & 11.22 & 0.200 & 0.793 & 14.10 & 0.290 & 0.647 & 12.89 & 0.326 & 0.652\\
average  & 13.06 & 0.318 & 0.576 & 13.07 & 0.243 & 0.580 & 14.17 & 0.318 & 0.578 & 15.29 & 0.367 & 0.585\\
\hline
\multicolumn{13}{c}{6 Views} \\
\hline
bicycle  & 14.35 & 0.188 & 0.576 & 12.92 & 0.181 & 0.663 & 15.76 & 0.294 & 0.597 & 16.52 & 0.311 & 0.604\\
bonsai  & 14.77 & 0.471 & 0.457 & 13.07 & 0.373 & 0.602 & 16.67 & 0.546 & 0.436 & 16.55 & 0.557 & 0.441\\
counter  & 15.09 & 0.428 & 0.467 & 13.77 & 0.352 & 0.535 & 16.02 & 0.495 & 0.449 & 16.99 & 0.545 & 0.428\\
garden  & 16.06 & 0.308 & 0.465 & 14.03 & 0.201 & 0.569 & 17.57 & 0.401& 0.504 & 18.74 & 0.406 & 0.490\\
room  & 14.80 & 0.481 & 0.446 & 13.98 & 0.426 & 0.564 & 15.22 & 0.542 & 0.443 & 17.54 & 0.623 & 0.410\\
stump  & 16.13 & 0.229 & 0.556 & 14.62 & 0.201 & 0.609 & 17.58 & 0.323 & 0.582 & 18.36 & 0.343 & 0.585 \\
kitchen  & 17.12 & 0.494 & 0.397 & 15.11 & 0.321 & 0.530 & 17.64 & 0.577 & 0.374 & 18.54 & 0.560 & 0.390\\
flowers  & 11.89 & 0.145 & 0.607 & 10.89 & 0.147 & 0.757 & 13.21 & 0.211 & 0.649 & 14.01 & 0.237 & 0.658\\
treehill & 13.33 & 0.240 & 0.584 & 12.10 & 0.222 & 0.741 & 15.46 & 0.347 & 0.613 & 15.36 & 0.363 & 0.605\\
average  & 14.96 & 0.355 & 0.505 & 15.02 & 0.338 & 0.506 & 16.12 & 0.415 & 0.517 & 17.16 & 0.447 & 0.500\\
\hline
\multicolumn{13}{c}{9 Views} \\
\hline
bicycle  & 15.30 & 0.237 & 0.536 & 13.53 & 0.213 & 0.648 & 17.15 & 0.343 & 0.577 & 17.10 & 0.332 & 0.578\\
bonsai  & 17.43 & 0.609 & 0.373 & 15.51 & 0.460 & 0.482 & 19.30 & 0.669 & 0.356 & 19.31 & 0.662 & 0.354\\
counter  & 16.42 & 0.516 & 0.406 & 14.54 & 0.391 & 0.493 & 17.63 & 0.572 & 0.391 & 18.23 & 0.607 & 0.379\\
garden  & 18.10 & 0.412 &  0.397 & 15.06 & 0.250 & 0.522 & 19.22 & 0.477& 0.455 & 19.97 & 0.470 & 0.446\\
room  & 17.36 & 0.600 & 0.370 & 15.49 & 0.492 & 0.499 & 18.16 & 0.662 & 0.359 & 19.75 & 0.700 & 0.366\\
stump  & 17.45 & 0.300 & 0.514 & 15.69 & 0.237 & 0.548 & 18.72 & 0.386 & 0.555 & 19.40 & 0.392 & 0.553 \\
kitchen  & 19.17 & 0.611 & 0.324 & 16.21 & 0.393 & 0.473 & 20.30 & 0.682 & 0.305 & 20.59 & 0.640 & 0.322\\
flowers  & 13.01 & 0.191 & 0.564 & 12.01 & 0.163 & 0.695 & 14.33 & 0.247 & 0.629 & 14.95 & 0.267 & 0.629\\
treehill & 14.34 & 0.300 & 0.555 & 13.23 & 0.265 & 0.733 & 15.46 & 0.347 & 0.613 & 15.98 & 0.390 & 0.595\\
average  & 16.79 & 0.447 & 0.446 & 16.67 & 0.423 & 0.449 & 17.94 & 0.492 & 0.471 & 18.36 & 0.496 & 0.465\\
\hline
    \end{tabular}
    \caption{Per-scene Quantitative comparison on sparse view reconstruction}
    \label{tab:per-scene}
\end{table*}

\end{document}